\documentclass[11pt]{article}

\usepackage{a4wide}
\usepackage{graphicx}
\usepackage[usenames]{color,colortbl}
\usepackage{pstricks,pst-node,pst-plot}
\usepackage{verbatim}
\usepackage{placeins}
\usepackage{array}
\usepackage{amstext,amsmath,amssymb}
\usepackage{booktabs}
\usepackage{lscape}
\usepackage{rotating}
\usepackage{subfigure}
\usepackage{algorithm}
\usepackage{algorithmic}
\usepackage{url}
\definecolor{lightgray}{gray}{.75}

\begin{document}

\title{On Solving the Oriented Two-Dimensional Bin Packing Problem under Free Guillotine Cutting: \\ Exploiting the Power of Probabilistic Solution Construction}

\author{Christian Blum$^1$ and Verena Schmid$^2$ and Lukas Baumgartner$^2$ \\
~\\
$^1$ALBCOM Research Group\\ 
Universitat Polit{\`e}cnica de Catalunya, Barcelona, Spain \\
{\sf cblum@lsi.upc.edu}\\
~\\
$^2$Department of Business Administration\\
Universit{\"a}t Wien, Vienna, Austria\\
{\sf \{verena.schmid,lukas.baumgartner\}@univie.ac.at}}

\date{}

\maketitle

\begin{abstract}
Two-dimensional bin packing problems are highly relevant combinatorial optimization problems. They find a large number of applications, for example, in the context of transportation or warehousing, and for the cutting of different materials such as glass, wood or metal. In this work we deal with the oriented two-dimensional bin packing problem under free guillotine cutting. In this specific problem a set of oriented rectangular items is given which must be packed into a minimum number of bins of equal size. The first algorithm proposed in this work is a randomized multi-start version of a constructive one-pass heuristic from the literature. Additionally we propose the use of this randomized one-pass heuristic within an evolutionary algorithm. The results of the two proposed algorithms are compared to the best approaches from the literature. In particular the evolutionary algorithm compares very favorably to current state-of-the-art approaches. The optimal solution for 4 previously unsolved instances could be found.
\end{abstract}

\section{Introduction}
\label{sec:intro}

Bin packing problems (BPPs) are well studied and highly popular combinatorial optimization problems. The main reason for their popularity is a large number of real-world applications. Moreover, in general they can be easily expressed in mathematical terms. In this work we deal with a specific variant of the two-dimensional bin packing problem (2BP), which consists in packing a set $\mathcal{Q} = \{1,\ldots,n\}$ of $n$ rectangular items into a minimum number of bins of height $H$ and width $W$ such that items do not overlap. Each item $j \in \mathcal{Q}$ is characterized by its height $h_j$ and its width $w_j$. Real world applications for the 2BP include, for example, cutting glass, wood or metal and packing in the context of transportation or warehousing (see ~\cite{HopTur:375,SwePat:691}). \\

According to Lodi et. al~\cite{LodMarVig:345} there are four different cases of the 2BP as described above. The differences between these four cases are based on two aspects: (1) a rotation of 90$^{\circ}$ of the items may, or may not, be permitted; (2) guillotine cutting may be required or free. The four resulting problem versions are as follows:
\begin{itemize}
	\item 2BP$\mid$O$\mid$G: Items are oriented and guillotine cutting is required.
	\item 2BP$\mid$O$\mid$F: Items are oriented and guillotine cuttings is free.
	\item 2BP$\mid$R$\mid$G: Items may be rotated by 90$^{\circ}$ and guillotine cutting is required.
	\item 2BP$\mid$R$\mid$F: Items may be rotated by 90$^{\circ}$ and guillotine cutting is free.
\end{itemize}

In this paper we exclusively focus on the 2BP$\mid$O$\mid$F version of the problem. Note that in the remainder of the paper the abbreviation 2BP will refer to this problem version. Concerning the complexity of the 2BP, Garey and Johnson classified the problem as NP-hard (see~\cite{GarJoh:book}).

\subsection{Existing Work}
\label{sec:existingsolutionmethods}

In general, different versions of the 2BP have been tackled in the literature by means of different integer programing models, heuristics, and exact algorithms. A good overview on the early work regarding the 2BP can be obtained from~\cite{LodMarVig:379,LodMarVig:241,Lodi:phd,DowDow:2}. In the following we will focus on existing heuristics as well as metaheuristics.

\subsection{Heuristics}

Concerning heuristics, the literature mainly distinguishes between \emph{one-phase} and \emph{two-phase approaches}. One-phase algorithms pack the items directly into the bins, whereas two-phase algorithms first pack the items into levels of one infinitely high strip with width $W$ and then stack these levels into the bins. Level-packing algorithms place items next to each other in each level. Hereby, the bottom of the first level is the bottom of the bin. For the next level the bottom is a horizontal line coinciding with the highest item of the level below. Note that, items can only be placed besides each other in each level, in contrast to packing items on top of each other. \\

Well known \emph{level-packing algorithms} are \textsc{Next-Fit Decreasing Height} (NFDH), \textsc{First-Fit Decreasing Height} (FFDH) and \textsc{Best-Fit Decreasing Height} (BFDH)~\cite{CofGarJohTar:808}. These strategies were originally developed for the one-dimensional bin packing problem, but have also been adapted to strip packing problems and for the application to the two-dimensional case. All three heuristics require the items to be sorted by non-increasing height, which represents the order in which they are packed. Moreover, they pack the items into one bin of infinite hight. \\

Next, \emph{two-phase level-packing algorithms} are shortly described. \textsc{Hybrid Next-Fit} (HNF) (see~\cite{FreGal:201}) is based on NFDH, \textsc{Hybrid First-Fit} (HFF)~\cite{ChuGarJoh:66} on FFDH and \textsc{Finite Best-Strip} (FBS)~\cite{BerWan:423}, which is also sometimes referred to as \textsc{Hybrid Best-Fit}, is based on BFDH. The first phase of all three algorithms consists in the execution of the heuristic on which they are based. This produces in each case a set of levels, which must then be packed into bins of finite hight. This is done by using the same strategy as the one that was used for the packing of the items into levels. Another example for a two-phase level-packing algorithm is \textsc{Knapsack Packing} (KP)~\cite{LodMarVig:345}. Phase one of KP consists in packing the levels by solving knapsack problems. Hereby, the tallest unpacked item, say $j$, initializes each new level. The remaining horizontal distance up to the right bin border ($W - w_j$) is taken as the capacity of the knapsack problem to be solved. Moreover, the width $w_i$ of any unpacked item $i$ is regarded as its weight, while the items' area $w_i \cdot h_i$ is regarded as its value (or profit). This procedure is repeated until all items are packed into levels. In the second phase the remaining one-dimensional bin packing problem is solved by using a heuristic such as \textsc{Best-Fit Decreasing} or an exact algorithm. Finally, \textsc{Floor Ceiling} (FC)~\cite{LodMarVig:345} can be seen as an improvement over FBS. Again, the first phase is used for packing items into levels, whereas these levels are packed into bins in the second phase. \\

Among the most important \emph{one-phase non-level-packing algorithms} are \textsc{Alternate Direction} (AD)~\cite{LodMarVig:345}, \textsc{Bottom-Left Fill} (BLF)~\cite{BakCofRiv:846}, \textsc{Improved Lowest Gap Fill} (LGFi) \cite{WonLee:334} and \textsc{Touching Perimeter} (TP)~\cite{LodMarVig:345}. In the following we describe these techniques shortly. AD sorts the items by non-increasing height and initializes $L$ bins, where $L$ is a lower bound for the necessary number of bins. Afterwards the bottom of the bins are filled from left to right using a best-fit decreasing strategy. Then one bin after another is being filled. In this context items are packed in bands from left to right and from right to left until no more items can be packed into the current bin. BLF initializes bins by placing the first item at the bottom left corner. The top left and bottom right corners of already placed items are positions at which the bottom left corner of new items may potentially be placed. BLF tries to place the items starting from the lowest to the highest available position. When positions with an equal height are encountered, the position closer to the left is tried first. LGFi has a preprocessing and a packing stage. In the preprocessing stage, items are sorted by non-increasing area as a first criterion. Ties are broken by non-increasing absolute difference between height and width of the items. The packing stage starts by initializing a bin with the first unpacked item, which is placed at the bottom left corner. Then items are placed at the bottom leftmost position. If possible, an item is chosen such that either the horizontal gap or the vertical gap is filled completely. If this is not possible, the largest fitting item is placed at this position. This is repeated until all items are packed. TP, the last one-phase non-level-packing algorithm considered here, first sorts the items by non-increasing area and initializes $L$ bins, where $L$ is a lower bound for the number of necessary bins. Furthermore, depending on a specific position in the bin, a score is associated to each item: the percentage of the edges of the item touching either an edge of another item or the border of the bin. Each item is now considered for different positions in the bin and for each of these positions the corresponding score is calculated. Each item is then placed at the position at which its score is highest. \\

The best heuristic for the 2BP which is currently available (labelled SCH) is based on solving a set-covering formulation of the problem~\cite{MonTot06:ijoc} by means of column generation. In the first phase, a rather small subset of all possible columns is generated by using greedy procedures and fast constructive heuristic algorithms from the literature. In the second phase, the resulting set-covering instance is solved by means of a Lagrangian-based heuristic. \\

In addition, some heuristics developed for three-dimensional packing can sometimes easily be applied to the 2BP. An example is the extreme point based heuristic from~\cite{CraPerTad:368}. This heuristic uses extreme points to determine all points in the bin where items can be placed. Extreme points can either be corners of the already placed items or points generated by the extended edges of the placed items. These points are updated every time an item is placed into the bin. For placing the items a modified version of BFDH is used.

\subsection{Metaheuristics}

The earliest metaheuristic developed for the 2BP is \emph{tabu search} (TS)~\cite{LodMarVig:158,LodMarVig:379}. An initial solution is created using a heuristic such as FBS, KP, or AD. Moreover, neighborhood moves are based on trying empty certain bins by repacking their items into other bins.

A metaheuristic based on \emph{guided local search} (GLS) has been presented in~\cite{FarEtAl03:ijoc}. This metaheuristic has its origins in constraint satisfaction applications. GLS uses memory to guide the search process away from already explored regions of the search space. This is done by adding a penalty term to the objective function that penalizes bad solution features of previously visited solutions.

A rather simple metaheuristic, labeled HBP, based on a greedy heuristic has been proposed in~\cite{BosMin03:4or}. HBP assigns a score to each item. Then, for the construction of a solution, the items are considered according to non-increasing values of the scores. After the construction of a solution the scores are updated using a certain criterion. This procedure is iterated until a pre-defined stopping criterion is met.

An approach labeled \emph{weight annealing} (WA) for solving the 2BP was proposed in~\cite{LohGolWas09:chapter}. The WA technique can be seen as an extension of a greedy heuristic. Hereby, weights are assigned to different parts of the solution space. These weights are changed during the execution of the algorithm on the basis of the generated solutions. Moreover, they have an influence on the decisions of the greedy heuristic when constructing a new solution.

Finally, the currently best-performing metaheuristic is a hybrid between a greedy randomized adaptive search procedure (GRASP) and variable neighborhood descent (VND)~\cite{ParEtAl10:aor}. The solution construction phase of GRASP is hereby based on a maximal-space heuristic from the field of container loading.

\subsection{Contribution of this Work}

In this paper we propose two algorithms based on a randomized version of the LGFi heuristic from the literature. First, a multi-start algorithm is developed. Second, our randomized version of LGFi is embedded into several operators of a comparatively simple evolutionary algorithm. Extensive computational experiments on publicly available benchmark instances show that both algorithms compare very favorably with the state of the art. In fact, the proposed multi-start algorithm and the evolutionary algorithm are able to solve 4 previously unsolved problem instances to optimality. Moreover, summing up the number of used bins concerning all 500 problem instances the evolutionary algorithm reaches a value of 7239, which is the best value reached by any algorithm that has been proposed for this problem.

\subsection{Organization of the Paper}

In Section~\ref{sec:problemformulation} we first outline an ILP model for the tackled problem. The proposed algorithms are then presented in Section~\ref{sec:solutionprocedure}. Finally, an experimental evaluation is provided in Section~\ref{sec:evaluation}, while conclusions and an outlook to the future are given in Section~\ref{sec:conclusions}.

\section{A New ILP Model}
\label{sec:problemformulation}

Inspired by the models proposed in~\cite{PisSig:36} and~\cite{PucRai:1304} we present in the following an alternative ILP model for the 2BP. For this purpose, we denote by $\mathcal{Q} = \{1,\ldots,n\}$ the set of all items and the set of all bins. $W$ and $H$ refer to the bin-width and the bin-height, while $w_{i}$ and $h_{i}$ refer to the width and the height of item $i \in \mathcal{Q}$. $W$, $H$, $w_{i}$ and $h_{i}$ are all integer values.

The binary decision variable $\alpha_{ik}$ evaluates to 1 if item $i$ is packed into bin $k$, and 0 otherwise. Only variables $\alpha_{ik}$ where $i \geq k$ are created so that only $\frac{n^2+n}{2}$ instead of $n^2$ have to be initialized. Furthermore items $\alpha_{kk}$ indicate if bins are opened or not. A bin is considered open if the item with the same index as the bin is placed in that bin. For example item 1 cannot be placed in bin 3 but only in bin 1. Item 3 can be placed in bin 3, in bin 2 in case item 2 is placed in bin 2, or in bin 1, which is always open as item 1 can only be placed in bin 1. It is easy to see that, even with this restricted variable set, all combinations of items packed into one bin are still possible. The integer variables $x_{i}$ and $y_{i}$ decide the x- and y-coordinates of each item within a bin. For the overlapping constraints, which we will introduce in the next paragraph, we need the binary variables $ul_{ij}, ua_{ij}, ur_{ij}$ and $uu_{ij}$. Each one of these four variables decides if item $i$ has to be to the left ($ul_{ij}$), above ($ua_{ij}$), to the right ($ur_{ij}$) or underneath ($uu_{ij}$) item $j$. Only variables for $i<j$ are created so that only $\frac{n^2-n}{2}$ instead of $n^2$ have to be initialized for each variable. This can be done because if item $i$ has to be to the left of item $j$, item $j$ automatically has to be to the right of item $i$ which makes it unnecessary to initialize the corresponding variable of item $j$.

{\allowdisplaybreaks
\begin{alignat}{2}
	& Z = \sum_{i=0}^n \alpha_{ii} \rightarrow min \label{fun:obj}\\
	& \sum_{k=0}^n \alpha_{ik} = 1 && i,k \in \mathcal{Q}; i \geq k \label{con:1}\\
	& \alpha_{ik} \leq \alpha_{kk} && i,k \in \mathcal{Q}; i \geq k \label{con:2}\\
	& x_{i} + w_{i} \leq W && i \in \mathcal{Q} \label{con:3}\\
	& y_{i} + h_{i} \leq H && i \in \mathcal{Q} \label{con:4}\\
	& ul_{ij} + ua_{ij} + ur_{ij} + uu_{ij} = 1 && i,j \in \mathcal{Q}; i < j \label{con:5}\\
	& x_{i} + w_{i} \leq x_{j} + W \cdot (3-ul_{ij}-\alpha_{ik}-\alpha_{jk}) \hspace{10 mm} && i,j,k \in \mathcal{Q}; k \leq i < j \label{con:6}\\
	& y_{i} + H \cdot (3-ua_{ij}-\alpha_{ik}-\alpha_{jk}) \geq y_{j} + h_{j} \hspace{10 mm} && i,j,k \in \mathcal{Q}; k \leq i < j \label{con:7}\\
	& x_{i} + W \cdot (3-ur_{ij}-\alpha_{ik}-\alpha_{jk}) \geq x_{j} + w_{j} \hspace{10 mm} && i,j,k \in \mathcal{Q}; k \leq i < j \label{con:8}\\
	& y_{i} + h_{i} \leq y_{j} + H \cdot (3-uu_{ij}-\alpha_{ik}-\alpha_{jk}) \hspace{10 mm} && i,j,k \in \mathcal{Q}; k \leq i < j \label{con:9}
\end{alignat}}

The objective function \eqref{fun:obj} minimizes the number of bins used. The constraint \eqref{con:1} ensures that each item is assigned to one and only one bin. That an item $i$ can only be assigned to an open/initialized bin is ensured by \eqref{con:2}. Constraints \eqref{con:3} and \eqref{con:4} ensure that each item is placed within the bin. Equation \eqref{con:5} states that item $i$ has to be placed either to the left, above, to the right or underneath item $j$. The last four equations \eqref{con:6}-\eqref{con:9} ensure that two items do not overlap if assigned to the same bin.

\section{The Proposed Algorithms}
\label{sec:solutionprocedure}

Both algorithms that we present in this paper are strongly based on heuristic LGFi, as developed by Wong and Lee in~\cite{WonLee:334}. LGFi itself is an improved version of the LGF heuristic presented by Lee in~\cite{Lee:34}. Note that LGFi is a two-stage heuristic. In the \emph{preprocessing stage} items are sorted into a list, while in the \emph{packing stage} these items are packed from the list into bins. More specifically, in the preprocessing stage items are sorted by non-increasing area as a first criterion. Ties are broken by non-increasing absolute difference between height and width of the items. The packing stage is an iterative process in which the following actions are performed at each iteration. First, the bottom leftmost position at which an item may be placed is identified. This position is henceforth called  the \emph{current position}. Then, two gaps are calculated with respect to this position. The \emph{horizontal gap} is defined as the distance between the current position and either the right border of the bin or the left edge of the first item between the current position and the right border of the bin. The distance between the current position and the upper border of the bin defines the value of the \emph{vertical gap}. The value of the smaller gap is called \emph{current gap}. The current gap is compared to either the widths of the items from the list of unpacked items, if the horizontal gap is the current gap, or to the heights of the items from the list of unpacked items, if the vertical gap is the current gap. The first item that fills the gap completely is placed with its bottom left corner at the identified position. If no such item exists, the first item which fits without any overlap is placed with its bottom left corner at the current position. If no such item exists either, some of the area must be declared \emph{wastage area}, which works as follows. A wastage area with the width of the horizontal gap is created. The height of the wastage area is chosen as the height of the upper edge of the lowest neighboring item, or, if no neighboring items exists, as the height of the bin. Finally, if no current position can be found, and if unpacked items exist, a new bin is opened.

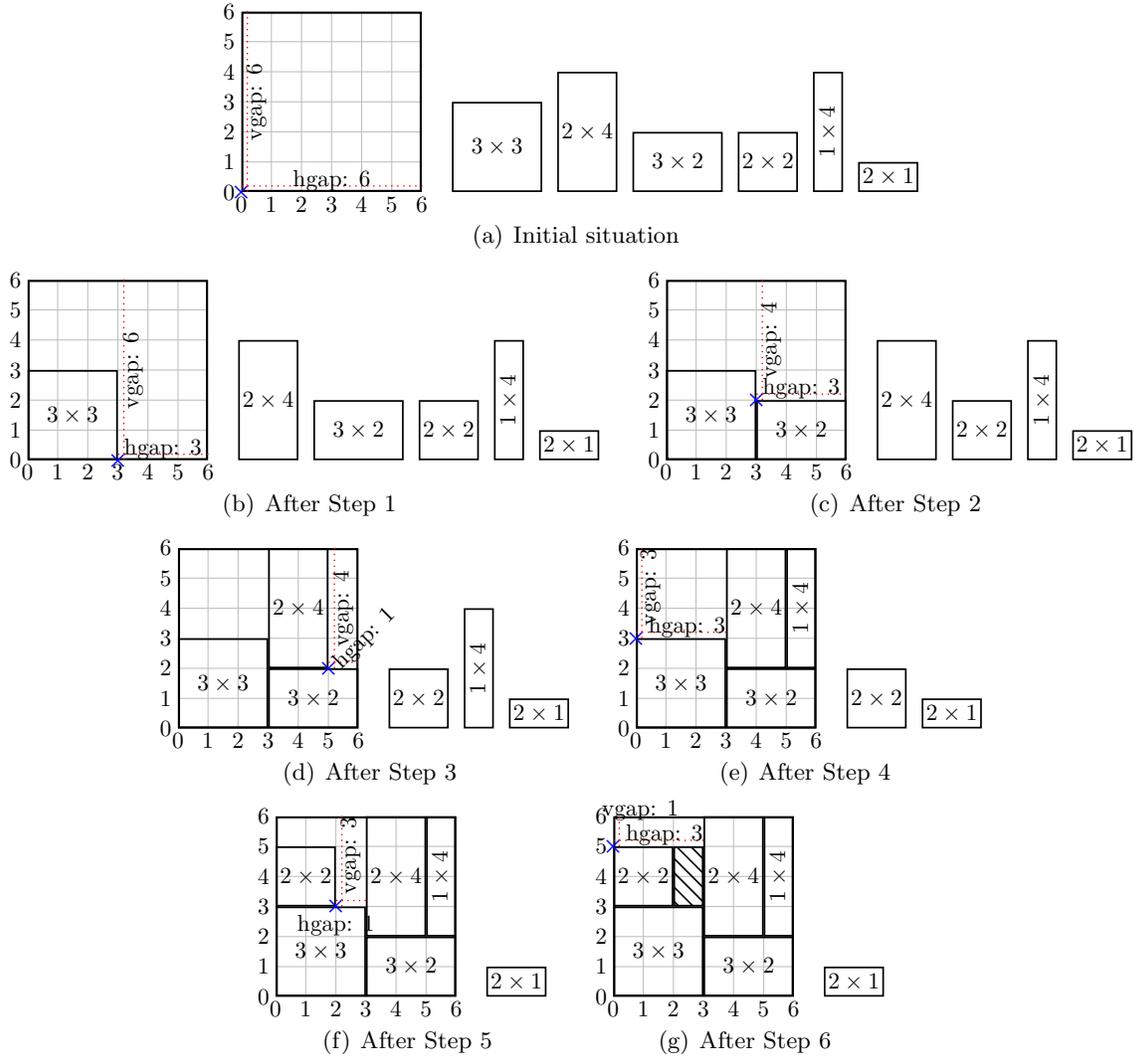
\begin{figure}[p]
\begin{center}
\subfigure[Initial situation]{
\label{fig:example:a}
\scalebox{0.8}{
\psset{xunit=.5mm,yunit=.5mm,runit=.5mm}
\begin{pspicture}(235,70)
\psline[linewidth=0.5pt,linecolor=lightgray]{-}(15,5)(15,65)
\psline[linewidth=0.5pt,linecolor=lightgray]{-}(25,5)(25,65)
\psline[linewidth=0.5pt,linecolor=lightgray]{-}(35,5)(35,65)
\psline[linewidth=0.5pt,linecolor=lightgray]{-}(45,5)(45,65)
\psline[linewidth=0.5pt,linecolor=lightgray]{-}(55,5)(55,65)
\psline[linewidth=0.5pt,linecolor=lightgray]{-}(5,15)(65,15)
\psline[linewidth=0.5pt,linecolor=lightgray]{-}(5,25)(65,25)
\psline[linewidth=0.5pt,linecolor=lightgray]{-}(5,35)(65,35)
\psline[linewidth=0.5pt,linecolor=lightgray]{-}(5,45)(65,45)
\psline[linewidth=0.5pt,linecolor=lightgray]{-}(5,55)(65,55)
\psframe[linewidth=1pt,fillstyle=none](5,5)(65,65)
\rput(5,1){0}
\rput(15,1){1}
\rput(25,1){2}
\rput(35,1){3}
\rput(45,1){4}
\rput(55,1){5}
\rput(65,1){6}
\rput(1,5){0}
\rput(1,15){1}
\rput(1,25){2}
\rput(1,35){3}
\rput(1,45){4}
\rput(1,55){5}
\rput(1,65){6}
\psdot[dotstyle=+,dotangle=45,dotsize=7pt,linecolor=blue](5,5)
\psline[linestyle=dotted,dotsep=2pt,linecolor=red]{-}(5,7)(65,7)
\rput(35,9){hgap: 6}
\psline[linestyle=dotted,dotsep=2pt,linecolor=red]{-}(7,5)(7,65)
\rput{90}(10,35){vgap: 6}
\psframe[linewidth=0.8pt,fillstyle=none](75,5)(105,35)
\rput(90,20){$3\times3$}
\psframe[linewidth=0.8pt,fillstyle=none](110,5)(130,45)
\rput(120,25){$2\times4$}
\psframe[linewidth=0.8pt,fillstyle=none](135,5)(165,25)
\rput(150,15){$3\times2$}
\psframe[linewidth=0.8pt,fillstyle=none](195,5)(205,45)
\rput{90}(200,25){$1\times4$}
\psframe[linewidth=0.8pt,fillstyle=none](170,5)(190,25)
\rput(180,15){$2\times2$}
\psframe[linewidth=0.8pt,fillstyle=none](210,5)(230,15)
\rput(220,10){$2\times1$}
\end{pspicture}}
}
\subfigure[After Step 1]{
\label{fig:example:b}
\scalebox{0.8}{
\psset{xunit=.5mm,yunit=.5mm,runit=.5mm}
\begin{pspicture}(200,70)
\psline[linewidth=0.5pt,linecolor=lightgray]{-}(15,5)(15,65)
\psline[linewidth=0.5pt,linecolor=lightgray]{-}(25,5)(25,65)
\psline[linewidth=0.5pt,linecolor=lightgray]{-}(35,5)(35,65)
\psline[linewidth=0.5pt,linecolor=lightgray]{-}(45,5)(45,65)
\psline[linewidth=0.5pt,linecolor=lightgray]{-}(55,5)(55,65)
\psline[linewidth=0.5pt,linecolor=lightgray]{-}(5,15)(65,15)
\psline[linewidth=0.5pt,linecolor=lightgray]{-}(5,25)(65,25)
\psline[linewidth=0.5pt,linecolor=lightgray]{-}(5,35)(65,35)
\psline[linewidth=0.5pt,linecolor=lightgray]{-}(5,45)(65,45)
\psline[linewidth=0.5pt,linecolor=lightgray]{-}(5,55)(65,55)
\psframe[linewidth=1pt,fillstyle=none](5,5)(65,65)
\rput(5,1){0}
\rput(15,1){1}
\rput(25,1){2}
\rput(35,1){3}
\rput(45,1){4}
\rput(55,1){5}
\rput(65,1){6}
\rput(1,5){0}
\rput(1,15){1}
\rput(1,25){2}
\rput(1,35){3}
\rput(1,45){4}
\rput(1,55){5}
\rput(1,65){6}
\psframe[linewidth=0.8pt,fillstyle=none](5,5)(35,35)
\rput(20,20){$3\times3$}
\psdot[dotstyle=+,dotangle=45,dotsize=7pt,linecolor=blue](35,5)
\psline[linestyle=dotted,dotsep=2pt,linecolor=red]{-}(35,7)(65,7)
\rput(50,9){hgap: 3}
\psline[linestyle=dotted,dotsep=2pt,linecolor=red]{-}(37,5)(37,65)
\rput{90}(40,35){vgap: 6}
\psframe[linewidth=0.8pt,fillstyle=none](75,5)(95,45)
\rput(85,25){$2\times4$}
\psframe[linewidth=0.8pt,fillstyle=none](100,5)(130,25)
\rput(115,15){$3\times2$}
\psframe[linewidth=0.8pt,fillstyle=none](160,5)(170,45)
\rput{90}(165,25){$1\times4$}
\psframe[linewidth=0.8pt,fillstyle=none](135,5)(155,25)
\rput(145,15){$2\times2$}
\psframe[linewidth=0.8pt,fillstyle=none](175,5)(195,15)
\rput(185,10){$2\times1$}
\end{pspicture}}
}
\subfigure[After Step 2]{
\label{fig:example:c}
\scalebox{0.8}{
\psset{xunit=.5mm,yunit=.5mm,runit=.5mm}
\begin{pspicture}(165,70)
\psline[linewidth=0.5pt,linecolor=lightgray]{-}(15,5)(15,65)
\psline[linewidth=0.5pt,linecolor=lightgray]{-}(25,5)(25,65)
\psline[linewidth=0.5pt,linecolor=lightgray]{-}(35,5)(35,65)
\psline[linewidth=0.5pt,linecolor=lightgray]{-}(45,5)(45,65)
\psline[linewidth=0.5pt,linecolor=lightgray]{-}(55,5)(55,65)
\psline[linewidth=0.5pt,linecolor=lightgray]{-}(5,15)(65,15)
\psline[linewidth=0.5pt,linecolor=lightgray]{-}(5,25)(65,25)
\psline[linewidth=0.5pt,linecolor=lightgray]{-}(5,35)(65,35)
\psline[linewidth=0.5pt,linecolor=lightgray]{-}(5,45)(65,45)
\psline[linewidth=0.5pt,linecolor=lightgray]{-}(5,55)(65,55)
\psframe[linewidth=1pt,fillstyle=none](5,5)(65,65)
\rput(5,1){0}
\rput(15,1){1}
\rput(25,1){2}
\rput(35,1){3}
\rput(45,1){4}
\rput(55,1){5}
\rput(65,1){6}
\rput(1,5){0}
\rput(1,15){1}
\rput(1,25){2}
\rput(1,35){3}
\rput(1,45){4}
\rput(1,55){5}
\rput(1,65){6}
\psframe[linewidth=0.8pt,fillstyle=none](5,5)(35,35)
\rput(20,20){$3\times3$}
\psframe[linewidth=0.8pt,fillstyle=none](35,5)(65,25)
\rput(50,15){$3\times2$}
\psdot[dotstyle=+,dotangle=45,dotsize=7pt,linecolor=blue](35,25)
\psline[linestyle=dotted,dotsep=2pt,linecolor=red]{-}(35,27)(65,27)
\rput(50,29){hgap: 3}
\psline[linestyle=dotted,dotsep=2pt,linecolor=red]{-}(37,25)(37,65)
\rput{90}(40,45){vgap: 4}
\psframe[linewidth=0.8pt,fillstyle=none](75,5)(95,45)
\rput(85,25){$2\times4$}
\psframe[linewidth=0.8pt,fillstyle=none](125,5)(135,45)
\rput{90}(130,25){$1\times4$}
\psframe[linewidth=0.8pt,fillstyle=none](100,5)(120,25)
\rput(110,15){$2\times2$}
\psframe[linewidth=0.8pt,fillstyle=none](140,5)(160,15)
\rput(150,10){$2\times1$}
\end{pspicture}}
}
\subfigure[After Step 3]{
\label{fig:example:d}
\scalebox{0.8}{
\psset{xunit=.5mm,yunit=.5mm,runit=.5mm}
\begin{pspicture}(140,70)
\psline[linewidth=0.5pt,linecolor=lightgray]{-}(15,5)(15,65)
\psline[linewidth=0.5pt,linecolor=lightgray]{-}(25,5)(25,65)
\psline[linewidth=0.5pt,linecolor=lightgray]{-}(35,5)(35,65)
\psline[linewidth=0.5pt,linecolor=lightgray]{-}(45,5)(45,65)
\psline[linewidth=0.5pt,linecolor=lightgray]{-}(55,5)(55,65)
\psline[linewidth=0.5pt,linecolor=lightgray]{-}(5,15)(65,15)
\psline[linewidth=0.5pt,linecolor=lightgray]{-}(5,25)(65,25)
\psline[linewidth=0.5pt,linecolor=lightgray]{-}(5,35)(65,35)
\psline[linewidth=0.5pt,linecolor=lightgray]{-}(5,45)(65,45)
\psline[linewidth=0.5pt,linecolor=lightgray]{-}(5,55)(65,55)
\psframe[linewidth=1pt,fillstyle=none](5,5)(65,65)
\rput(5,1){0}
\rput(15,1){1}
\rput(25,1){2}
\rput(35,1){3}
\rput(45,1){4}
\rput(55,1){5}
\rput(65,1){6}
\rput(1,5){0}
\rput(1,15){1}
\rput(1,25){2}
\rput(1,35){3}
\rput(1,45){4}
\rput(1,55){5}
\rput(1,65){6}
\psframe[linewidth=0.8pt,fillstyle=none](5,5)(35,35)
\rput(20,20){$3\times3$}
\psframe[linewidth=0.8pt,fillstyle=none](35,5)(65,25)
\rput(50,15){$3\times2$}
\psframe[linewidth=0.8pt,fillstyle=none](35,25)(55,65)
\rput(45,45){$2\times4$}
\psdot[dotstyle=+,dotangle=45,dotsize=7pt,linecolor=blue](55,25)
\psline[linestyle=dotted,dotsep=2pt,linecolor=red]{-}(55,27)(65,27)
\rput{45}(67,35){hgap: 1}
\psline[linestyle=dotted,dotsep=2pt,linecolor=red]{-}(57,25)(57,65)
\rput{90}(60,45){vgap: 4}
\psframe[linewidth=0.8pt,fillstyle=none](100,5)(110,45)
\rput{90}(105,25){$1\times4$}
\psframe[linewidth=0.8pt,fillstyle=none](75,5)(95,25)
\rput(85,15){$2\times2$}
\psframe[linewidth=0.8pt,fillstyle=none](115,5)(135,15)
\rput(125,10){$2\times1$}
\end{pspicture}}
}
\subfigure[After Step 4]{
\label{fig:example:e}
\scalebox{0.8}{
\psset{xunit=.5mm,yunit=.5mm,runit=.5mm}
\begin{pspicture}(125,70)
\psline[linewidth=0.5pt,linecolor=lightgray]{-}(15,5)(15,65)
\psline[linewidth=0.5pt,linecolor=lightgray]{-}(25,5)(25,65)
\psline[linewidth=0.5pt,linecolor=lightgray]{-}(35,5)(35,65)
\psline[linewidth=0.5pt,linecolor=lightgray]{-}(45,5)(45,65)
\psline[linewidth=0.5pt,linecolor=lightgray]{-}(55,5)(55,65)
\psline[linewidth=0.5pt,linecolor=lightgray]{-}(5,15)(65,15)
\psline[linewidth=0.5pt,linecolor=lightgray]{-}(5,25)(65,25)
\psline[linewidth=0.5pt,linecolor=lightgray]{-}(5,35)(65,35)
\psline[linewidth=0.5pt,linecolor=lightgray]{-}(5,45)(65,45)
\psline[linewidth=0.5pt,linecolor=lightgray]{-}(5,55)(65,55)
\psframe[linewidth=1pt,fillstyle=none](5,5)(65,65)
\rput(5,1){0}
\rput(15,1){1}
\rput(25,1){2}
\rput(35,1){3}
\rput(45,1){4}
\rput(55,1){5}
\rput(65,1){6}
\rput(1,5){0}
\rput(1,15){1}
\rput(1,25){2}
\rput(1,35){3}
\rput(1,45){4}
\rput(1,55){5}
\rput(1,65){6}
\psframe[linewidth=0.8pt,fillstyle=none](5,5)(35,35)
\rput(20,20){$3\times3$}
\psframe[linewidth=0.8pt,fillstyle=none](35,5)(65,25)
\rput(50,15){$3\times2$}
\psframe[linewidth=0.8pt,fillstyle=none](35,25)(55,65)
\rput(45,45){$2\times4$}
\psframe[linewidth=0.8pt,fillstyle=none](55,25)(65,65)
\rput{90}(60,45){$1\times4$}
\psdot[dotstyle=+,dotangle=45,dotsize=7pt,linecolor=blue](5,35)
\psline[linestyle=dotted,dotsep=2pt,linecolor=red]{-}(5,37)(35,37)
\rput(22,39){hgap: 3}
\psline[linestyle=dotted,dotsep=2pt,linecolor=red]{-}(7,35)(7,65)
\rput{90}(10,52){vgap: 3}
\psframe[linewidth=0.8pt,fillstyle=none](75,5)(95,25)
\rput(85,15){$2\times2$}
\psframe[linewidth=0.8pt,fillstyle=none](100,5)(120,15)
\rput(110,10){$2\times1$}
\end{pspicture}}
}
\subfigure[After Step 5]{
\label{fig:example:f}
\scalebox{0.8}{
\psset{xunit=.5mm,yunit=.5mm,runit=.5mm}
\begin{pspicture}(100,70)
\psline[linewidth=0.5pt,linecolor=lightgray]{-}(15,5)(15,65)
\psline[linewidth=0.5pt,linecolor=lightgray]{-}(25,5)(25,65)
\psline[linewidth=0.5pt,linecolor=lightgray]{-}(35,5)(35,65)
\psline[linewidth=0.5pt,linecolor=lightgray]{-}(45,5)(45,65)
\psline[linewidth=0.5pt,linecolor=lightgray]{-}(55,5)(55,65)
\psline[linewidth=0.5pt,linecolor=lightgray]{-}(5,15)(65,15)
\psline[linewidth=0.5pt,linecolor=lightgray]{-}(5,25)(65,25)
\psline[linewidth=0.5pt,linecolor=lightgray]{-}(5,35)(65,35)
\psline[linewidth=0.5pt,linecolor=lightgray]{-}(5,45)(65,45)
\psline[linewidth=0.5pt,linecolor=lightgray]{-}(5,55)(65,55)
\psframe[linewidth=1pt,fillstyle=none](5,5)(65,65)
\rput(5,1){0}
\rput(15,1){1}
\rput(25,1){2}
\rput(35,1){3}
\rput(45,1){4}
\rput(55,1){5}
\rput(65,1){6}
\rput(1,5){0}
\rput(1,15){1}
\rput(1,25){2}
\rput(1,35){3}
\rput(1,45){4}
\rput(1,55){5}
\rput(1,65){6}
\psframe[linewidth=0.8pt,fillstyle=none](5,5)(35,35)
\rput(20,20){$3\times3$}
\psframe[linewidth=0.8pt,fillstyle=none](35,5)(65,25)
\rput(50,15){$3\times2$}
\psframe[linewidth=0.8pt,fillstyle=none](35,25)(55,65)
\rput(45,45){$2\times4$}
\psframe[linewidth=0.8pt,fillstyle=none](55,25)(65,65)
\rput{90}(60,45){$1\times4$}
\psframe[linewidth=0.8pt,fillstyle=none](5,35)(25,55)
\rput(15,45){$2\times2$}
\psdot[dotstyle=+,dotangle=45,dotsize=7pt,linecolor=blue](25,35)
\psline[linestyle=dotted,dotsep=2pt,linecolor=red]{-}(25,37)(35,37)
\rput(25,29){hgap: 1}
\psline[linestyle=dotted,dotsep=2pt,linecolor=red]{-}(27,35)(27,65)
\rput{90}(30,52){vgap: 3}
\psframe[linewidth=0.8pt,fillstyle=none](75,5)(95,15)
\rput(85,10){$2\times1$}
\end{pspicture}}
}
\subfigure[After Step 6]{
\label{fig:example:g}
\scalebox{0.8}{
\psset{xunit=.5mm,yunit=.5mm,runit=.5mm}
\begin{pspicture}(100,70)
\psline[linewidth=0.5pt,linecolor=lightgray]{-}(15,5)(15,65)
\psline[linewidth=0.5pt,linecolor=lightgray]{-}(25,5)(25,65)
\psline[linewidth=0.5pt,linecolor=lightgray]{-}(35,5)(35,65)
\psline[linewidth=0.5pt,linecolor=lightgray]{-}(45,5)(45,65)
\psline[linewidth=0.5pt,linecolor=lightgray]{-}(55,5)(55,65)
\psline[linewidth=0.5pt,linecolor=lightgray]{-}(5,15)(65,15)
\psline[linewidth=0.5pt,linecolor=lightgray]{-}(5,25)(65,25)
\psline[linewidth=0.5pt,linecolor=lightgray]{-}(5,35)(65,35)
\psline[linewidth=0.5pt,linecolor=lightgray]{-}(5,45)(65,45)
\psline[linewidth=0.5pt,linecolor=lightgray]{-}(5,55)(65,55)
\psframe[linewidth=1pt,fillstyle=none](5,5)(65,65)
\rput(5,1){0}
\rput(15,1){1}
\rput(25,1){2}
\rput(35,1){3}
\rput(45,1){4}
\rput(55,1){5}
\rput(65,1){6}
\rput(1,5){0}
\rput(1,15){1}
\rput(1,25){2}
\rput(1,35){3}
\rput(1,45){4}
\rput(1,55){5}
\rput(1,65){6}
\psframe[linewidth=0.8pt,fillstyle=none](5,5)(35,35)
\rput(20,20){$3\times3$}
\psframe[linewidth=0.8pt,fillstyle=none](35,5)(65,25)
\rput(50,15){$3\times2$}
\psframe[linewidth=0.8pt,fillstyle=none](35,25)(55,65)
\rput(45,45){$2\times4$}
\psframe[linewidth=0.8pt,fillstyle=none](55,25)(65,65)
\rput{90}(60,45){$1\times4$}
\psframe[linewidth=0.8pt,fillstyle=none](5,35)(25,55)
\rput(15,45){$2\times2$}
\psframe[linewidth=0.8pt,fillstyle=vlines](25,35)(35,55)
\psdot[dotstyle=+,dotangle=45,dotsize=7pt,linecolor=blue](5,55)
\psline[linestyle=dotted,dotsep=2pt,linecolor=red]{-}(5,57)(35,57)
\rput(22,59){hgap: 3}
\psline[linestyle=dotted,dotsep=2pt,linecolor=red]{-}(7,55)(7,65)
\rput(14,67){vgap: 1}
\psframe[linewidth=0.8pt,fillstyle=none](75,5)(95,15)
\rput(85,10){$2\times1$}
\end{pspicture}}
}
\end{center}
\caption{Example of the working of LFGi. (a) shows the initial situtation. Each subsequent subfigure shows the situation after placing one more item. In each subfigure the remaining items are shown to the right of the bin, ordered from left to right. The last step, which is not shown, consists in placing the last remaining item at position $(0,5)$.}
\label{fig:example}
\end{figure}

\paragraph{Example.} Figure~\ref{fig:example} shows the working of LGFi by means of a simple example. The left-hand side of each graphic shows the bin which is currently packed. The cross marks the current position, while the dotted lines show the horizontal and the vertical gap (indicated by hgap and vgap). The unpacked items are shown sorted from left to right at the right-hand side of each graphic. The body of each item shows its dimensions. In the initial situation (see Figure~\ref{fig:example:a}), the current postion corresponds to the bottom left corner of the empty bin. As the current gap evaluates to 6, no item is able to fill the current gap completely. Therefore, the first item from the list is chosen and placed at the current position. After this first step (see Figure~\ref{fig:example:b}), the current position is $(3,0)$, and the current gap (as defined by the horizontal gap) evaluates to 3. The first item from the list which fills this gap completely is the second item (with dimensions $3 \times 2$). Therefore, this item is chosen and placed at the current position. The packing stage of LGFi proceeds in the same way until reaching the situation shown in Figure~\ref{fig:example:f}. The current position at this point is $(2,3)$, and the current gap (corresponding to the horizontal gap) evaluates to 1. Unfortunately, the only remaining unpacked item does not fit without overlap at this position. Therefore, a wastage area must be declared. The width of this wastage space is equal to the horizontal gap. The height of the wastage space is 2, because after two vertical space units, the upper border of the neighboring item to the left is reached. Finally, as a last step, the last unpacked item is placed at position $(0,5)$.

\subsection{Multistart LGFi}
\label{sec:mslgfi}

The main idea of this paper is the use of the LGFi heuristic in a probabilistic way within the preprocessing stage. Our first approach is described in the following. As mentioned before, the preprocessing stage of LGFi generates an input sequence of all items. In this input sequence, items are ordered with respect to non-increasing area. In the following, $pos_i$ refers to the position of an item $i$ in this sequence. Multistart LGFi (MS-LGFi) works as follows. At each iteration, a new input sequence $s$ is probabilistically generated on the basis of the original input sequence. Then, this new input sequence is provided to LGFi for the generation of the packing. At the end of the algorithm, the best found solution is provided as output.

In the following we explain the way in which a new input sequence $s$ is generated based on the original input sequence. Remember that the total number of items is denoted by $n$. A value $v_i$ is then assigned to each item $i$ in the following way:
\begin{equation}
  v_i := (n - pos_i)^{\kappa}
\end{equation}
where $\kappa \geq 1$ is a parameter. The positions of $s$ are filled from 1 to $n$ in an iterative way. At each step, let $\mathcal{I} \subseteq \mathcal{Q}$ be the set of items that are not yet assigned to $s$. An item $i \in \mathcal{I}$ is chosen according to probabilities $\mathbf{p}(i|\mathcal{I})$ (for all $i \in \mathcal{I}$) by roulette-wheel-selection. These probabilities $\mathbf{p}(i|\mathcal{I})$ are calculated proportional to $v_i$:
\begin{equation}
  \mathbf{p}(i|\mathcal{I}) = \frac{v_{i}}{ \sum_{i \in \mathcal{I}} v_{i} }
\end{equation}
Note that the larger parameter $\kappa$, the more similar the newly generated input sequence $s$ will be to the original deterministic sequence.

\subsection{Evolutionary Algorithm}

The MS-LGFi algorithm, as proposed in the previous subsection, may have the disadvantage that no learning takes place over time. In other words, MS-LGFi may only find good input sequences for LGFi by chance. Moreover, once a good input sequence has been found, the knowledge about this sequence is forgotten at the end of the corresponding iteration. Therefore, we started to investigate if, for example, an evolutionary algorithm would is able learn good input sequences for LGFi. For this purpose the following evolutionary algorithm for the 2BP---henceforth labeled EA-LGFi---was devised.

A solution in the context of EA-LGFi is an input sequence $s$ for LGFi. Note that $s$ is an ordered list of all items that must be packed. The item at position $j$ of this list (where $j=1,\ldots,n$) is denoted by $s_j$. The function value $f(s)$ of a solution $s$ is calculated by applying LGFi to $s$. The pseudo-code of EA-LGFi is shown in Alg.~\ref{algo:ea}. The first step of EA-LGFi consists in generating the initial population of size $p_{\mbox{\small size}}$ (see function $\textsf{GenerateInitialPopulation}(p_{\mbox{\small size}},\kappa)$). Then, at each iteration a crossover operator is applied in function $\textsf{Crossover}(P,c_{\mbox{\small rate}},\delta)$, recreating $c_{\mbox{\small rate}}$ percent of the population. This provides a population $P'$ with less than $p_{\mbox{\small size}}$ solutions. The missing $p_{\mbox{\small size}}-|P'|$ solutions are generated by function $\textsf{AddNewSolutions}(P',p_{\mbox{\small size}},\kappa)$. In the following the three functions of algorithm EA-LGFi are outlined in more detail.

\begin{algorithm}[t!]
\caption{Evolutionary Algorithm for the 2BP (EA-LGFi)}
\label{algo:ea}
\begin{algorithmic}[1]
    \STATE {\bf input:} values for parameters $p_{\mbox{\small size}}$, $c_{\mbox{\small rate}}$, $\kappa$ and $\delta$
    \STATE $P := \textsf{GenerateInitialPopulation}(p_{\mbox{\small size}},\kappa)$
    \WHILE{stopping criterion not met}
         \STATE $P' := \textsf{Crossover}(P,c_{\mbox{\small rate}},\delta)$
         \STATE $P := \textsf{AddNewSolutions}(P',p_{\mbox{\small size}},\kappa)$
    \ENDWHILE
    \STATE {\bf output:} best solution found
\end{algorithmic}
\end{algorithm}

\paragraph{$\textsf{GenerateInitialPopulation}(p_{\mbox{\small size}},\kappa)$} In this function, $p_{\mbox{\small size}}$ solutions are probabilistically generated in the same way as in MS-LGFi (see Section~\ref{sec:mslgfi}). Parameter $\kappa$ is used for this purpose.

\paragraph{$\textsf{Crossover}(P,c_{\mbox{\small rate}},\delta)$} This operator applies recombination to each of the best $\lfloor c_{\mbox{\small rate}} \cdot |P| \rfloor$ solutions of $P$, where $0 < c_{\mbox{\small rate}} \leq 1$ is a parameter of the algorithm. Extensive empirical tests have shown that a crossover rate $c_{\mbox{\small rate}} = 0.7$ works best for the instances at hand. For each solution $s$ from the set of best $\lfloor c_{\mbox{\small rate}} \cdot |P| \rfloor$ solutions of $P$, a crossover parter $s^c \in P$ (such that $s^c \not= s$) is chosen from $P$ by means of roulette-wheel-selection. Assume that $P$ is an ordered list in which solutions are sorted according to their objective function values in a non-increasing manner. Ties are broken by the load of the last bin, that is, solutions with a lower load in the last bin are ordered first. Let $pos(s)$ denote the position of a solution $s$ in $P$. The probability $\mathbf{p}(s^c|s)$ for a solution $s^c \not= s$ to be chosen as a crossover partner for solution $s \in P$ is as follows:
\begin{equation}
  \mathbf{p}(s^c|s) := \frac{(p_{\mbox{\small size}} - 1 - pos(s^c))^{\delta}}{ \sum_{s^o \in P, s^o \not= s} (p_{\mbox{\small size}} - 1 - pos(s^o))^{\delta}} \enspace
\end{equation}
where $\delta \geq 1$ is a parameter of the algorithm. Given two crossover partners $s$ and $s^c$, one offspring solution $s^{\mbox{\small off}}$ is generated as explained in the following. First, three pointers ($k$, $l$ and $r$) are initialized to the first position. Then, the $n$ positions of $s^{\mbox{\small off}}$ are filled from 1 to $n$ as follows. If $s_k = s^c_l$ then $s^{\mbox{\small off}}_r := s_k$. In words, if position $k$ of solution $s$ and position $l$ of solution $s^c$ contain the same item, then this item is placed at position $r$ of the offspring solution $s^{\mbox{\small off}}$. Next, position pointer $r$ is incremented, and position pointers $k$ and $l$ are moved to the right until reaching the closest position containing an item which does not yet appear in solution $s^{\mbox{\small off}}$. In case $s_k \not= s^c_l$, the item for position $r$ of solution $s^{\mbox{\small off}}$ is chosen probabilistically among $s_k$ and $s^c_l$, where a probability of $0.75$ is given to the item originating from the better of the two solutions. Afterwards, the position pointer $r$ is incremented. Moreover, the position pointer of the solution from which the item was selected is moved to the right until reaching the closest position containing an item which does not yet appear in solution $s^{\mbox{\small off}}$. The resulting solution $s^{\mbox{\small off}}$ is evaluated by using it as input for LGFi. In case $f(s^{\mbox{\small off}}) < f(s)$ or $f(s^{\mbox{\small off}}) = f(s)$ and $s^{\mbox{\small off}}$ has a lower load than $s$ in the last bin, solution $s^{\mbox{\small off}}$ is added to the new population $P'$, otherwise solution $s$ is added to $P'$.

\paragraph{$\textsf{AddNewSolutions}(P',p_{\mbox{\small size}},\kappa)$} This function probabilistically generates $p_{\mbox{\small size}} - |P'|$ solutions in the same way as in MS-LGFi (see Section~\ref{sec:mslgfi}). Parameter $\kappa$ is used for this purpose.

\section{Experimental Evaluation}
\label{sec:evaluation}

MS-LGFi and EA-LGFi were implemented in ANSI C++ using GCC 4.4 for compiling the software. The experimental results that we outline in the following were obtained on a PC with an AMD64X2 4400 processor and 4 Gigabyte of memory. The proposed algorithms were applied to a benchmark set of 500 problem instances from the literature. After an initial study of the algorithms' behavior, a detailed experimental evaluation is presented.

\subsection{Problem Instances}
\label{subsec:instances}

Ten classes of problem instances for the 2BP are provided in the literature. A first instance set, containing six classes (I-VI), was proposed by Berkey and Wang in~\cite{BerWan:423}. For each of these classes, the widths and heights of the items were chosen uniformly at random from the intervals presented in Table~\ref{tab:classesivi}. Moreover, the classes differ in the width ($W$) and the height ($H$) of the bins. Instance sizes, in terms of the number of items, are taken from $\{20,40,60,80,100\}$. Berkey and Wang provided 10 instances for each combination of a class with an instance size. This results in a total of 300 problem instances.

\begin{table}[!ht]
  \centering
  \caption{Specification of instance classes I-VI (as provided by~\cite{BerWan:423}).}
    \begin{tabular}{ccccc}
    \toprule
    Class & $w_{j}$    & $h_{j}$    & $W$     & $H$ \\
    \midrule
    I     & [1,10] & [1,10] & 10    & 10 \\
    II    & [1,10] & [1,10] & 30    & 30 \\
    III   & [1,35] & [1,35] & 40    & 40 \\
    IV    & [1,35] & [1,35] & 100   & 100 \\
    V     & [1,100] & [1,100] & 100   & 100 \\
    VI    & [1,100] & [1,100] & 300   & 300 \\
    \bottomrule
    \end{tabular}
  \label{tab:classesivi}
\end{table}

The second instance set, consisting of classes VII-X, was introduced by Martello and Vigo in~\cite{MarVig:388}. In general, they considered four different types of items, as presented in Table~\ref{tab:types}. The four item types differ in the limits for the width $w_i$ and the height $h_i$ of an item. Then, based on these four item types, Martello and Vigo introduced four classes of instances which differ in the percentage of items they contain from each type. As an example, let us consider an instance of class VII. $70\%$ of the items of such an instance are of type 1, $10\%$ of the items are of type 2, further $10\%$ of the items are of type 3, and the remaining $10\%$ of the items are of type 4. These percentages are given per class in Table~\ref{tab:classesviix}. As in the case of the first instance set, instance sizes are taken from $\{20,40,60,80,100\}$. The instance set by Martello and Vigo consists of 10 instances for each combination of a class with an instance size. This results in a total of 200 problem instances.

\begin{table}[!ht]
  \centering
  \caption{Item types for classes VII-X (as introduced in~\cite{MarVig:388}).}
    \begin{tabular}{ccccc}
    \toprule
    Item type  & $w_{j}$    & $h_{j}$    & W     & H \\
    \midrule
    1     & $[\frac{2}{3} \cdot W,W]$     & $[1,\frac{1}{2} \cdot H]$     & 100   & 100 \\
    2     & $[1,\frac{1}{2} \cdot W]$     & $[\frac{2}{3} \cdot H,H]$     & 100   & 100 \\
    3     & $[\frac{1}{2} \cdot W,W]$     & $[\frac{1}{2} \cdot H,H]$     & 100   & 100 \\
    4     & $[1,\frac{1}{2} \cdot W]$     & $[1,\frac{1}{2} \cdot H]$     & 100   & 100 \\
    \bottomrule
    \end{tabular}
  \label{tab:types}
\end{table}

\begin{table}[!ht]
  \centering
  \caption{Specification of instance classes VII-X (as provided by~\cite{MarVig:388}).}
    \begin{tabular}{ccccc}
    \toprule
    Class & Type 1 & Type 2 & Type 3 & Type 4 \\
    \midrule
    VII   & 70\%  & 10\%  & 10\%  & 10\% \\
    VIII  & 10\%  & 70\%  & 10\%  & 10\% \\
    IX    & 10\%  & 10\%  & 70\%  & 10\% \\
    X     & 10\%  & 10\%  & 10\%  & 70\% \\
    \bottomrule
    \end{tabular}
  \label{tab:classesviix}
\end{table}

These 500 instances can be downloaded from \url{http://www.or.deis.unibo.it/research.html}.

\subsection{Algorithm Tuning}

\begin{figure}[!t]
\begin{center}
  \includegraphics[width=7cm,angle=270]{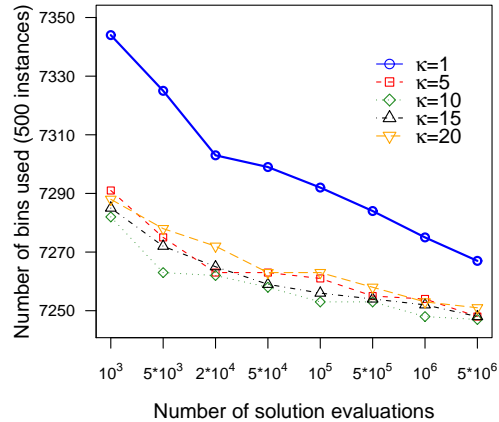}
\end{center}
\caption{Tuning results for MS-LGFi.}
\label{fig:ms-tuning}
\end{figure}

In order to study the behavior of MS-LGFi we performed tests with a varying limit for the number of solution evaluations (that is, algorithm iterations) and for various settings of parameter $\kappa$. More specifically, we considered limits for the number of solution evaluations from $\{10^3,5\cdot10^3,2\cdot10^4,5\cdot10^4,10^5,5\cdot10^5,10^6,5\cdot10^6\}$ and values for $\kappa$ from $\{1,5,10,15,20\}$. For each combination of the two parameters MS-LGFi was applied exactly once to each of the 500 problem instances. The sum of the number of bins used in the best solutions generated for all 500 problem instances is used as a measure. The graphic in Figure~\ref{fig:ms-tuning} provides this information for all parameter value combinations. The best performance is generally achieved (for each solution evaluation limit) with the setting $\kappa = 10$. Moreover, when increasing the number of solution evaluations from $10^6$ to $5 \cdot 10^6$, the algorithm performance improves only slightly. Therefore, the final results of MS-LGFi that are presented in the following section are obtained with $\kappa = 10$ and a number of $5 \cdot 10^6$ solutions evaluations. \\

\begin{figure}[!t]
\begin{center}
  \includegraphics[width=7cm,angle=270]{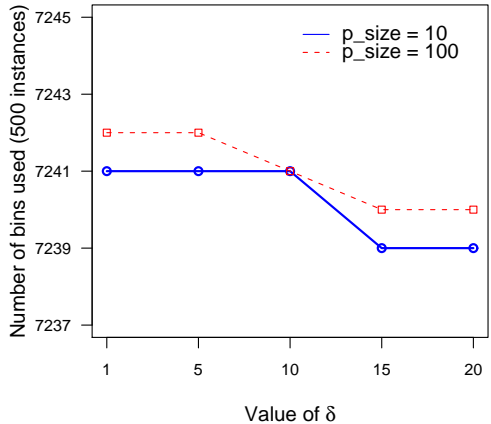}
\end{center}
\caption{Tuning results for EA-LGFi.}
\label{fig:ea-tuning}
\end{figure}

Concerning EA-LGFi, the same number of solution evaluations as for MS-LGFi was chosen as a stopping criterion (that is, $5 \cdot 10^6$ solution evaluations). Moreover, for parameter $\kappa$ we chose value 10, as in the case of MS-LGFi. However, we tested different population sizes ($p_{\mbox{\small size}} \in \{10,100\}$) and different values for parameter $\delta$ ($\delta \in \{1,5,10,15,20\}$). Remember that the value of $\delta$ is used for the calculation of the probabilities for solutions to be selected as crossover partners. In general, the higher the value of $\delta$, the more are good solutions preferred over worse ones, when selecting a crossover partner $s^c$ for $s$. EA-LGFi was applied for each combination of $\delta$ and $p_{\mbox{\small size}} \in \{10,100\}$ exactly once to each of the 500 problem instances. The sum of the number of bins used in the best solutions generated for all 500 instances is shown in Figure~\ref{fig:ea-tuning} for each parameter value combination. Even though differences in algorithm performance are quite small, higher values of $\delta$ seem to work better than smaller ones. Moreover, a population size of 10 generally seems to work slightly better than a population size of 100. The final results of EA-LGFi presented in the following section are the ones obtained with $\delta = 20$ and $p_{\mbox{\small size}} = 10$.

\subsection{Numerical Results}

The following six benchmark algorithms were chosen from the literature: The set covering heuristic (SCH) from~\cite{MonTot06:ijoc}, the hybrid GRASP approach from~\cite{ParEtAl10:aor}, the HBP approach from~\cite{BosMin03:4or}, the tabu search (TS) algorithm from~\cite{LodMarVig:158,LodMarVig:379}, the guided local search (GLS) approach from~\cite{FarEtAl03:ijoc}, and the weighted annealing (WA) metaheuristic from~\cite{LohGolWas09:chapter}. Among these approaches, SCH and GRASP are currently regarded to be the state-of-the-art techniques for the 2BP.

The results are shown in Tables~\ref{tab:comparison-1} and~\ref{tab:comparison-2} in a way which is traditional for the 2BP. For each algorithm the results are shown in two columns. The first column (with heading {\bf value}) provides the sum of the number of bins used in the best solutions generated for the 10 instances of a combination between instance class (I -- X) and number of items (20 -- 100). For example, the best solutions generated for the 10 instances of Class~I (20 items) by algorithm SCH occupy in total 71 bins. In case a value corresponds to the best result obtained by any algorithm, it highlighted in bold. Moreover, in the case of MS-LGFi and EA-LGFi a value is marked by an asterisk if it is better than the best know value as of today. The second column (with heading {\bf time (s)}) shows the average computation time (in seconds) necessary to find the best solutions for the 10 instances of a combination between instance class and number of items. For example, algorithm SCH needed on average 0.06 seconds to find its best solutions for the 10 instances of Class I (20 items). The only exception is GLS for which the second column is missing, as the computation time information was not given in the original paper. Finally, the last line of Table~\ref{tab:comparison-2} provides a summary of the results over all 500 problem instances. For each algorithm is given the sum of number of bins used, as well as the average computation time, for the 500 instances.

There are several aspects about the results that should be mentioned. First, the number bins used by the best solutions generated by EA-LGFi for the 500 problem instances amounts to 7239, which is the best value ever achieved by any algorithm. The best algorithm so far (GRASP) achieved a value of 7241. Moreover, MS-LGFi achieves a value of 7247, which is the 3rd best value ever obtained by any algorithm. Only GRASP (7241) and SCH (7243) achieve better values. This is remarkable, because MS-LGFi is a simple multi-start algorithm. It is also interesting to note that both MS-LGFi and EA-LGFi are able to solve four problem instances to optimality that have never been solved before. This concerns instances 173 and 174 (both from Class IV, with 60 items), instance 197 (from Class IV, with 100 items), and instance 298 (from Class VI, with 100 items). Detailed results for each single instance are shown in the 10 tables of Appendix~A.

Finally, we would like to comment on the computation times. Due to the fact that different processors and different computation time limits have been used for the generation of the results, the computation times are certainly not directly comparable. However, the computation times of all algorithms are, in general, very low. Therefore, no algorithm can be identified to have a particular advantage or disadvantage over the other algorithms for what concerns the computation time requirements. In the following we provide the information about processors and computation time limits for the competitor algorithms: SCH was run on a Digital Alpha 533 MHz with a time limit of 100 seconds per instance. The same machine and time limit was used for HBP, because the results presented in Tables~\ref{tab:comparison-1} and~\ref{tab:comparison-2} are the ones from a re-implementation from~\cite{MonTot06:ijoc}. GRASP was executed on a Pentium Mobile with 1500 MHz with a stopping criterion of 50000 iterations per application. Furthermore, TS was tested on a Silicon Graphics INDY R10000sc with 195 MHz and a computation time limit of 60 seconds per problem instance. Finally, GLS was executed on a Digital 500au workstation with a 500 MHz 21164 CPU using a computation time limit of 100 seconds per problem instance, while WA was run on a Pentium 4 with 3 GHz.

\begin{sidewaystable}[p]
  \centering
  \caption{Part A: Numerical results for the 250 instances of the first 5 instance classes (Class~I -- Class~V).}
		\scalebox{0.8}{
    \begin{tabular}{c|cccccccccccc|cc|cc}
    \addlinespace
    \toprule
& {\bf LB} & \multicolumn{2}{c}{\bf SCH} & \multicolumn{2}{c}{\bf GRASP} & \multicolumn{2}{c}{\bf HBP} & \multicolumn{2}{c}{\bf TS} & \multicolumn{1}{c}{\bf GLS} & \multicolumn{2}{c|}{\bf WA} & \multicolumn{2}{c}{\bf MS-LGFi} & \multicolumn{2}{|c}{\bf EA-LGFi} \\ \cmidrule(r){3-4}\cmidrule(r){5-6}\cmidrule(r){7-8}\cmidrule(r){9-10}\cmidrule(r){12-13}
&          & value & time (s) & value & time (s) & value & time (s) & value & time (s) & value & value & time (s) & value & time (s) & value & time (s) \\ \midrule
Class I    &     &     &       &     &       &     &        &     &       &     &     &       &      &       &      &       \\
      20   &  71 & {\bf 71} &  0.06 & {\bf 71} &  0.00 & {\bf 71} &  10.09 &  {\bf 71} & 24.00 &  {\bf 71} &  {\bf 71} &  0.21 & {\bf 71}   &  0.00 & {\bf 71}	&  0.00 \\
      40   & 134 & {\bf 134} &  2.42 & {\bf 134} &  0.00 & {\bf 134} &  32.02 & 135 & 36.11 & {\bf 134} & {\bf 134} &  0.06 & {\bf 134}  &  0.00 & {\bf 134}  &  0.00 \\
      60   & 197 & {\bf 200} &  7.26 & {\bf 200} &  4.50 & 201 &  40.17 & 201 & 48.93 & 201 & {\bf 200} &  0.67 & {\bf 200}  &  0.00 & {\bf 200}  &  0.01 \\
      80   & 274 & {\bf 275} &  4.63 & {\bf 275} &  1.50 & {\bf 275} &  10.10 & 282 & 48.17 & {\bf 275} & {\bf 275} &  3.07 & {\bf 275}  &  0.00 & {\bf 275}  &  0.00 \\
     100   & 317 & {\bf 317} &  5.21 & {\bf 317} &  0.00 & 319 &  20.79 & 326 & 60.81 & 321 & {\bf 317} &  9.21 & {\bf 317}  &  0.00 & {\bf 317}  &  0.00 \\ \midrule
Class II   &	 &     &       &     &	    &	  &	   &	 &	 &     &     &       &      &	    &	   &	   \\
      20   &  10 &  {\bf 10} &  0.06 &  {\bf 10} &  0.00 &  {\bf 10} &	0.06 &  {\bf 10} &  0.01 &  {\bf 10} &  {\bf 10} &  0.05 & {\bf 10}   &  0.00 & {\bf 10}   &  0.00 \\
      40   &  19 &  {\bf 19} &  0.67 &  {\bf 19} &  0.00 &  {\bf 19} &	1.33 &  20 &  0.01 &  {\bf 19} &  20 &  0.04 & {\bf 19}   &  0.00 & {\bf 19}   &  0.00 \\
      60   &  25 &  {\bf 25} &  0.07 &  {\bf 25} &  0.00 &  {\bf 25} &	0.07 &  27 &  0.09 &  {\bf 25} &  {\bf 25} &  0.43 & {\bf 25}	&  0.00 & {\bf 25}   &  0.00 \\
      80   &  31 &  {\bf 31} &  0.07 &  {\bf 31} &  0.00 &  {\bf 31} &	1.35 &  33 & 12.00 &  32 &  {\bf 31} & 13.89 & {\bf 31}   &  0.00 & {\bf 31}   &  0.00 \\
     100   &  39 &  {\bf 39} &  0.79 &  {\bf 39} &  0.00 &  {\bf 39} &	0.26 &  40 &  6.00 &  {\bf 39} &  {\bf 39} &  8.70 & {\bf 39}	&  0.00 & {\bf 39}   &  0.00 \\ \midrule
Class III  &	 &     &       &     &	    &	  &	   &	 &	 &     &     &       &      &	    &	   &	   \\
      20   &  51 &  {\bf 51} &  0.07 &  {\bf 51} &  0.00 &  {\bf 51} &  20.74 &  55 & 54.00 &  {\bf 51} &  53 &  0.04 & {\bf 51}   &  0.00 & {\bf 51}   &  0.02 \\
      40   &  92 &  {\bf 94} &  2.66 &  {\bf 94} &  3.00 &  {\bf 94} &  21.38 &  97 & 54.02 &  95 &  {\bf 94} &  2.15 & {\bf 94}   &  0.02 & {\bf 94}   &  0.01 \\
      60   & 136 & {\bf 139} &  6.21 & {\bf 139} &  4.60 & 140 &  40.19 & 140 & 45.67 & 140 & {\bf 139} &  0.16 & {\bf 139}  &  0.03 & {\bf 139}  &  0.27 \\
      80   & 187 & {\bf 189} &  8.80 & {\bf 189} &  4.10 & 190 &  32.72 & 198 & 54.31 & 193 & {\bf 189} &  3.16 & 191  &  0.31 & {\bf 189}  & 20.68 \\
     100   & 221 & {\bf 223} & 12.80 & {\bf 223} &  4.90 & 225 &  41.51 & 236 & 60.10 & 229 & 224 &  7.52 & 225  &  1.03 & 224  & 26.17 \\ \midrule
Class IV   &	 &     &       &     &	    &	  &	   &	 &	 &     &     &       &      &	    &	   &	   \\
      20   &  10 &  {\bf 10} &  0.06 &  {\bf 10} &  0.00 &  {\bf 10} &	0.07 &  {\bf 10} &  0.01 &  {\bf 10} &  {\bf 10} &  0.05 & {\bf 10}   &  0.00 & {\bf 10}   &  0.00 \\
      40   &  19 &  {\bf 19} &  0.07 &  {\bf 19} &  0.00 &  {\bf 19} &	0.08 &  {\bf 19} &  0.01 &  {\bf 19} &  {\bf 19} &  0.14 & {\bf 19}   &  0.00 & {\bf 19}   &  0.00 \\
      60   &  23 &  25 &  6.15 &  25 &  3.00 &  25 &  20.15 &  26 &  0.14 &  25 &  25 &  0.05 & {\bf 23$^{\ast}$}   & 11.77 & {\bf 23$^{\ast}$}   & 12.18 \\
      80   &  30 &  32 & 10.35 &  {\bf 31} &  1.90 &  32 &  21.67 &  33 & 18.00 &  33 &  {\bf 31} & 12.19 & {\bf 31}   &  0.00 & {\bf 31}   &  0.00 \\
     100   &  37 &  38 &  4.72 &  38 &  1.50 &  38 &  12.02 &  38 &  6.00 &  39 &  38 &  0.36 & {\bf 37$^{\ast}$}   &  0.01 & {\bf 37$^{\ast}$}   &  0.00 \\ \midrule
Class V    &	 &     &       &     &	    &	  &	   &	 &	 &     &     &       &      &	    &	   &	   \\
      20   &  65 &  {\bf 65} &  0.06 &  {\bf 65} &  0.00 &  {\bf 65} &   0.10 &  66 & 36.02 &  {\bf 65} &  {\bf 65} &  0.10 & {\bf 65}   &  0.00 & {\bf 65}   &  0.00 \\
      40   & 119 & {\bf 119} &  1.98 & {\bf 119} &  0.00 & {\bf 119} &  30.78 & {\bf 119} & 27.07 & {\bf 119} & {\bf 119} &  0.17 & {\bf 119}  &  0.05 & {\bf 119}  &  0.03 \\
      60   & 179 & {\bf 180} &  1.93 & {\bf 180} &  1.50 & {\bf 180} &  27.07 & 182 & 56.77 & 181 & {\bf 180} &  1.49 & {\bf 180}  &  0.07 & {\bf 180}  &  0.14 \\
      80   & 241 & {\bf 247} & 20.66 & {\bf 247} &  9.00 & 248 &  62.19 & 251 & 56.18 & 250 & {\bf 247} &  2.66 & {\bf 247}  &  0.09 & {\bf 247}  &  0.03 \\
     100   & 279 & {\bf 282} & 18.50 & {\bf 282} &  5.20 & 286 &  61.07 & 295 & 60.34 & 288 & 283 &  3.50 & 287  &  1.53 & 284  & 27.33 \\
\bottomrule
\end{tabular}}
\label{tab:comparison-1}
\end{sidewaystable}

\begin{sidewaystable}[p]
  \centering
  \caption{Part B: Numerical results for the 250 instances of the last 5 instance classes (Class~VI -- Class~X). The last table row provides a summary of the results for all 10 instances classes.}
		\scalebox{0.8}{
    \begin{tabular}{c|cccccccccccc|cc|cc}
    \addlinespace
    \toprule
& {\bf LB} & \multicolumn{2}{c}{\bf SCH} & \multicolumn{2}{c}{\bf GRASP} & \multicolumn{2}{c}{\bf HBP} & \multicolumn{2}{c}{\bf TS} & \multicolumn{1}{c}{\bf GLS} & \multicolumn{2}{c|}{\bf WA} & \multicolumn{2}{c}{\bf MS-LGFi} & \multicolumn{2}{|c}{\bf EA-LGFi} \\ \cmidrule(r){3-4}\cmidrule(r){5-6}\cmidrule(r){7-8}\cmidrule(r){9-10}\cmidrule(r){12-13}
&          & value & time (s) & value & time (s) & value & time (s) & value & time (s) & value & value & time (s) & value & time (s) & value & time (s) \\ \midrule
Class VI   &     &     &       &     &       &     &        &     &       &     &     &       &      &       &      &       \\
      20   &  10 &  {\bf 10} &  0.06 &  {\bf 10} &  0.00 &  {\bf 10} &   0.07 &  {\bf 10} &  0.01 &  {\bf 10} &  {\bf 10} &  0.04 &  {\bf 10} &  0.00 &  {\bf 10} &  0.00 \\
      40   &  15 &  {\bf 17} &  6.85 &  {\bf 17} &  3.00 &  {\bf 17} &  22.69 &  19 &  0.03 &  18 &  19 &  0.07 &  {\bf 17} &  0.01 &  {\bf 17}  &  0.03 \\
      60   &  21 &  {\bf 21} &  0.66 &  {\bf 21} &  0.10 &  {\bf 21} &   0.16 &  22 &  0.04 &  22 &  22 &  0.05 &  {\bf 21} &  0.00 &  {\bf 21}  &  0.00 \\
      80   &  30 &  {\bf 30} &  0.23 &  {\bf 30} &  0.00 &  {\bf 30} &   0.23 &  {\bf 30} &  0.01 &  {\bf 30} &  {\bf 30} &  0.06 &  {\bf 30} &  0.00 &  {\bf 30}  &  0.00 \\
     100   &  32 &  34 &  6.29 &  34 &  3.00 &  34 &  20.42 &  34 & 12.00 &  34 &  33 & 21.00 &  {\bf 32$^{\ast}$} & 29.57 &  {\bf 32$^{\ast}$}  &  0.58 \\ \midrule
Class VII  &	 &     &       &     &	    &	  &	   &	 &	 &     &     &       &      &	    &	   &	   \\
      20   &  55 &  {\bf 55} &  0.13 &  {\bf 55} &  0.00 &  {\bf 55} & 20.12 &  {\bf 55} & 12.02 &  {\bf 55} &  {\bf 55} &  0.05 &  {\bf 55} &  0.00 &  {\bf 55}   &  0.00 \\
      40   & 109 & {\bf 111} &  3.02 & {\bf 111} &  3.00 & 112 & 33.56 & 114 & 37.01 & 113 & {\bf 111} &  2.12 & {\bf 111} &  0.01 & {\bf 111}   &  0.01 \\
      60   & 156 & {\bf 158} &  8.85 & 159 &  4.50 & 160 & 43.33 & 162 & 36.44 & 159 & 159 &  6.79 & 159 &  0.00 & 159   &  0.00 \\
      80   & 224 & {\bf 232} & 54.79 & {\bf 232} & 12.00 & {\bf 232} & 80.35 & {\bf 232} & 54.52 & {\bf 232} & {\bf 232} &  0.27 & {\bf 232} &  0.00 & {\bf 232}   &  0.00 \\
     100   & 269 & {\bf 271} & 25.06 & {\bf 271} &  3.10 & 273 & 42.82 & 277 & 47.43 & 275 & {\bf 271} &  1.92 & {\bf 271} &  0.53 & {\bf 271}   &  0.01 \\ \midrule
Class VIII &	 &     &       &     &	     &	   &  &	   &	 &     &     &       &      &	    &	   &	   \\
      20   &  58 &  {\bf 58} &  0.06 &  {\bf 58} &  0.00 &  {\bf 58} &   0.07 &  {\bf 58} & 18.04 &  {\bf 58} &  {\bf 58} &  0.05 &  {\bf 58} &  0.01 & {\bf 58}   &  0.03 \\
      40   & 112 & {\bf 113} &  0.96 & {\bf 113} &  1.50 & {\bf 113} &  11.36 & 114 & 18.72 & 114 & {\bf 113} &  0.21 & {\bf 113} &  0.01 & {\bf 113}   &  0.00 \\
      60   & 159 & 162 &  9.05 & {\bf 161} &  4.20 & 162 &  30.81 & 162 & 20.99 & 163 & 162 &  0.16 & {\bf 161} &  0.00 & {\bf 161}  &  0.02 \\
      80   & 223 & {\bf 224} & 11.60 & {\bf 224} &  1.60 & 225 &  20.83 & 226 & 37.95 & 225 & {\bf 224} &  0.33 & {\bf 224} &  0.01 & {\bf 224}  &  0.00 \\
     100   & 274 & 279 & 47.13 & 278 &  6.10 & 279 &  50.98 & 284 & 52.66 & 281 & {\bf 277} &  0.06 & 278 &  0.05 & {\bf 277}  &  0.25 \\ \midrule
Class IX   &	 &     &       &     &	    &	  &	   &	 &	 &     &     &       &      &	    &	   &	   \\
      20   & 143 & {\bf 143} &  0.06 & {\bf 143} &  0.00 & {\bf 143} &  0.06 & {\bf 143} &  0.01 & {\bf 143} & {\bf 143} &  0.19 & {\bf 143} &  0.00 & {\bf 143}   &  0.00 \\
      40   & 278 & {\bf 278} &  0.07 & {\bf 278} &  0.00 & {\bf 278} &  0.07 & {\bf 278} & 24.05 & {\bf 278} & 279 &  0.04 & {\bf 278} &  0.00 & {\bf 278}   &  0.00 \\
      60   & 437 & {\bf 437} &  0.07 & {\bf 437} &  0.10 & {\bf 437} &  0.07 & 438 & 24.26 & {\bf 437} & 438 &  0.12 & {\bf 437} &  0.00 & {\bf 437}   &  0.00 \\
      80   & 577 & {\bf 577} &  0.08 & {\bf 577} &  0.00 & {\bf 577} &  0.09 & {\bf 577} & 54.31 & {\bf 577} & {\bf 577} &  0.16 & {\bf 577} &  0.00 & {\bf 577}   &  0.00 \\
     100   & 695 & {\bf 695} &  0.11 & {\bf 695} &  0.00 & {\bf 695} &  0.11 & {\bf 695} & 34.11 & {\bf 695} & {\bf 695} &  0.23 & {\bf 695} &  0.00 & {\bf 695}   &  0.00 \\ \midrule
Class X    &	 &     &       &     &	    &	  &	   &	 &	 &     &     &       &      &	    &	   &	   \\
      20   &  42 &  {\bf 42} &  0.12 &  {\bf 42} &  0.00 &  {\bf 42} &  15.73 &  43 & 12.00 &  {\bf 42} &  43 &  0.29 &  {\bf 42} &  0.05 & {\bf 42}   &  0.02 \\
      40   &  74 &  {\bf 74} &  0.11 &  {\bf 74} &  0.00 &  {\bf 74} &  20.14 &  75 & 25.18 &  {\bf 74} &  {\bf 74} &  0.21 &  {\bf 74} &  0.00 & {\bf 74}  &  0.00 \\
      60   &  98 & 101 &  8.89 & {\bf 100} &  4.50 & 102 &  53.39 & 104 & 42.13 & 102 & 102 &  0.16 & 101 &  2.61 & 101  &  0.71 \\
      80   & 123 & {\bf 128} & 38.26 & 129 &  9.40 & 130 &  70.35 & 130 & 47.30 & 130 & 129 &  5.42 & 129 &  3.15 & {\bf 128}  &  0.06 \\
     100   & 153 & {\bf 159} & 55.77 & {\bf 159} &  9.20 & 160 &  88.00 & 166 & 60.10 & 162 & {\bf 159} &  9.26 & 160 &  0.14 & 160  &  0.08 \\ \midrule
{\bf Summary} & 7173 & 7243 & 7.90 & 7241 & 2.21 & 7265 & 22.67 & 7358 & 28.72 & 7284 & 7253 & 2.39 & 7247 & 1.02 & {\bf 7239} & 1.77 \\
\bottomrule
\end{tabular}}
\label{tab:comparison-2}
\end{sidewaystable}

\section{Conclusions and Outlook}
\label{sec:conclusions}

In this paper we presented two algorithms for tackling the oriented two-dimensional bin packing problem under free guillotine cutting (2BP). Both algorithms are strongly based on a probabilistic version of an existing one-pass heuristic (LGFi) from the literature. The first algorithm is a simple multistart metaheuristics, whereas the second one is an evolutionary algorithm. The results have shown that both algorithms obtain very good results in comparison to current state-of-the-art approaches. In fact, both algorithms are able to solve four problem instances---which have not been solved yet by any algorithm---to optimality. Moreover, the best solutions generated by the evolutionary algorithm for the 500 instances use, in total, a number 7239 bins. This is the best value ever achieved by any algorithm proposed for the 2BP.

In the future we plan to investigate additional ways in which the probabilistic version of LGFi might be exploited. For example, an ant colony optimization approach might be better suited than an evolutionary algorithm for learning input sequences for LGFi. Moreover, we plan to add a local search procedure to our algorithms for improving the constructed solutions.

\section*{Acknowledgements}

This work was supported by the binational grant \emph{Acciones Integradas} ES16-2009 (Austria) and MEC HA2008-0005 (Spain), and by grant TIN2007-66523 (FORMALISM) of the Spanish government. In addition, Christian Blum acknowledges support from the \textit{Ram{\'o}n y Cajal} program of the Spanish Government of which he is a research fellow.

\newpage

\section*{Appendix~A}

This appendix contains 10 tables, one for the 50 problem instances of each instance class. Each table provides the results of MS-LGFi and EA-LGFi for each instance. The structure of the tables is as follows. The first column contains the number of items. The second column provides the instance number (numbered from 1 to 500). The next two columns contain information about the currently best known lower and upper bound values for each instance. Finally, the results of MS-LGFi, as well as the results of EA-LGFi, are presented in three columns. The first one of these three columns (with heading {\bf res}) provides the number of bins used in the best found solution. In case a value in this column is shown with a gray background, the upper bound for the corresponding problem instances was improved. On the other side, in case a value is shown within a frame of white background, the best known upper bound for the corresponding instance was not reached. The second column (with heading {\bf eval}) indicates after how many solution evaluations the best solution was found, while the third column (with heading {\bf time}) provides the computation time (in seconds) after which the best solution was found.

\begin{table}[!t]
\centering
\caption{Detailed results for the 50 problem instances of Class~I.}
\label{tab:detailed-results-class1}
\scalebox{0.8}{
\begin{tabular}{ll|ll|lll|lll} \hline
{\bf \#}    &            &          &          & \multicolumn{3}{|c|}{\bf MS-LGFi} & \multicolumn{3}{|c}{\bf EA-LGFi} \\
{\bf items} & {\bf inst} & {\bf LB} & {\bf UB} & {\bf res} & {\bf eval} & {\bf time} & {\bf res} & {\bf eval} & {\bf time} \\ \hline
20 & 1 & 8 & 8 & 8 & 1 & 0.0 & 8 & 1 & 0.0 \\
   & 2 & 5 & 5 & 5 & 2 & 0.0 & 5 & 1 & 0.0 \\
   & 3 & 9 & 9 & 9 & 1 & 0.0 & 9 & 1 & 0.0 \\
   & 4 & 6 & 6 & 6 & 1 & 0.0 & 6 & 1 & 0.0 \\
   & 5 & 6 & 6 & 6 & 327 & 0.0 & 6 & 360 & 0.0 \\
   & 6 & 9 & 9 & 9 & 1 & 0.0 & 9 & 1 & 0.0 \\
   & 7 & 6 & 6 & 6 & 1 & 0.0 & 6 & 1 & 0.0 \\
   & 8 & 6 & 6 & 6 & 29 & 0.0 & 6 & 68 & 0.0 \\
   & 9 & 8 & 8 & 8 & 1 & 0.0 & 8 & 1 & 0.0 \\
   & 10 & 8 & 8 & 8 & 1 & 0.0 & 8 & 1 & 0.0 \\ \hline
40 & 11 & 10 & 10 & 10 & 1 & 0.0 & 10 & 1 & 0.0 \\
   & 12 & 12 & 12 & 12 & 1 & 0.0 & 12 & 1 & 0.0 \\
   & 13 & 17 & 17 & 17 & 1 & 0.0 & 17 & 1 & 0.0 \\
   & 14 & 14 & 14 & 14 & 5 & 0.0 & 14 & 2 & 0.0 \\
   & 15 & 15 & 15 & 15 & 3 & 0.0 & 15 & 1 & 0.0 \\
   & 16 & 14 & 14 & 14 & 1 & 0.0 & 14 & 1 & 0.0 \\
   & 17 & 11 & 11 & 11 & 511 & 0.0 & 11 & 1220 & 0.0 \\
   & 18 & 19 & 19 & 19 & 1 & 0.0 & 19 & 1 & 0.0 \\
   & 19 & 11 & 11 & 11 & 1 & 0.0 & 11 & 3 & 0.0 \\
   & 20 & 11 & 11 & 11 & 1 & 0.0 & 11 & 1 & 0.0 \\ \hline
60 & 21 & 22 & 23 & 23 & 2 & 0.0 & 23 & 1 & 0.0 \\
   & 22 & 18 & 19 & 19 & 1 & 0.0 & 19 & 1 & 0.0 \\
   & 23 & 21 & 21 & 21 & 1 & 0.0 & 21 & 2 & 0.0 \\
   & 24 & 22 & 22 & 22 & 1 & 0.0 & 22 & 1 & 0.0 \\
   & 25 & 19 & 19 & 19 & 1 & 0.0 & 19 & 1 & 0.0 \\
   & 26 & 17 & 17 & 17 & 63 & 0.0 & 17 & 2465 & 0.1 \\
   & 27 & 15 & 16 & 16 & 1 & 0.0 & 16 & 1 & 0.0 \\
   & 28 & 21 & 21 & 21 & 7 & 0.0 & 21 & 10 & 0.0 \\
   & 29 & 18 & 18 & 18 & 1 & 0.0 & 18 & 9 & 0.0 \\
   & 30 & 24 & 24 & 24 & 1 & 0.0 & 24 & 1 & 0.0 \\ \hline
80 & 31 & 24 & 25 & 25 & 1 & 0.0 & 25 & 1 & 0.0 \\
   & 32 & 26 & 26 & 26 & 1 & 0.0 & 26 & 1 & 0.0 \\
   & 33 & 27 & 27 & 27 & 1 & 0.0 & 27 & 1 & 0.0 \\
   & 34 & 27 & 27 & 27 & 2 & 0.0 & 27 & 2 & 0.0 \\
   & 35 & 26 & 26 & 26 & 1 & 0.0 & 26 & 2 & 0.0 \\
   & 36 & 28 & 28 & 28 & 4 & 0.0 & 28 & 67 & 0.0 \\
   & 37 & 31 & 31 & 31 & 1 & 0.0 & 31 & 1 & 0.0 \\
   & 38 & 29 & 29 & 29 & 1 & 0.0 & 29 & 1 & 0.0 \\
   & 39 & 30 & 30 & 30 & 1 & 0.0 & 30 & 2 & 0.0 \\
   & 40 & 26 & 26 & 26 & 1 & 0.0 & 26 & 4 & 0.0 \\ \hline
100 & 41 & 28 & 28 & 28 & 3 & 0.0 & 28 & 3 & 0.0 \\
   & 42 & 31 & 31 & 31 & 24 & 0.0 & 31 & 116 & 0.0 \\
   & 43 & 29 & 29 & 29 & 1 & 0.0 & 29 & 1 & 0.0 \\
   & 44 & 30 & 30 & 30 & 11 & 0.0 & 30 & 103 & 0.0 \\
   & 45 & 32 & 32 & 32 & 1 & 0.0 & 32 & 2 & 0.0 \\
   & 46 & 37 & 37 & 37 & 1 & 0.0 & 37 & 1 & 0.0 \\
   & 47 & 28 & 28 & 28 & 6 & 0.0 & 28 & 23 & 0.0 \\
   & 48 & 33 & 33 & 33 & 1 & 0.0 & 33 & 5 & 0.0 \\
   & 49 & 31 & 31 & 31 & 227 & 0.0 & 31 & 181 & 0.0 \\
   & 50 & 38 & 38 & 38 & 2 & 0.0 & 38 & 1 & 0.0 \\ \hline
\end{tabular}}
\end{table}

\begin{table}[!t]
\centering
\caption{Detailed results for the 50 problem instances of Class~II.}
\label{tab:detailed-results-class2}
\scalebox{0.8}{
\begin{tabular}{ll|ll|lll|lll} \hline
{\bf \#}    &            &          &          & \multicolumn{3}{|c|}{\bf MS-LGFi} & \multicolumn{3}{|c}{\bf EA-LGFi} \\
{\bf items} & {\bf inst} & {\bf LB} & {\bf UB} & {\bf res} & {\bf eval} & {\bf time} & {\bf res} & {\bf eval} & {\bf time} \\ \hline
20 & 51 & 1 & 1 & 1 & 1 & 0.0 & 1 & 1 & 0.0 \\
   & 52 & 1 & 1 & 1 & 1 & 0.0 & 1 & 1 & 0.0 \\
   & 53 & 1 & 1 & 1 & 1 & 0.0 & 1 & 1 & 0.0 \\
   & 54 & 1 & 1 & 1 & 1 & 0.0 & 1 & 1 & 0.0 \\
   & 55 & 1 & 1 & 1 & 1 & 0.0 & 1 & 1 & 0.0 \\
   & 56 & 1 & 1 & 1 & 1 & 0.0 & 1 & 1 & 0.0 \\
   & 57 & 1 & 1 & 1 & 1 & 0.0 & 1 & 1 & 0.0 \\
   & 58 & 1 & 1 & 1 & 1 & 0.0 & 1 & 1 & 0.0 \\
   & 59 & 1 & 1 & 1 & 1 & 0.0 & 1 & 1 & 0.0 \\
   & 60 & 1 & 1 & 1 & 1 & 0.0 & 1 & 1 & 0.0 \\ \hline
40 & 61 & 1 & 1 & 1 & 166 & 0.0 & 1 & 81 & 0.0 \\
   & 62 & 2 & 2 & 2 & 1 & 0.0 & 2 & 1 & 0.0 \\
   & 63 & 2 & 2 & 2 & 1 & 0.0 & 2 & 1 & 0.0 \\
   & 64 & 2 & 2 & 2 & 1 & 0.0 & 2 & 1 & 0.0 \\
   & 65 & 2 & 2 & 2 & 1 & 0.0 & 2 & 1 & 0.0 \\
   & 66 & 2 & 2 & 2 & 1 & 0.0 & 2 & 1 & 0.0 \\
   & 67 & 2 & 2 & 2 & 1 & 0.0 & 2 & 1 & 0.0 \\
   & 68 & 2 & 2 & 2 & 1 & 0.0 & 2 & 1 & 0.0 \\
   & 69 & 2 & 2 & 2 & 1 & 0.0 & 2 & 1 & 0.0 \\
   & 70 & 2 & 2 & 2 & 1 & 0.0 & 2 & 1 & 0.0 \\ \hline
60 & 71 & 3 & 3 & 3 & 1 & 0.0 & 3 & 1 & 0.0 \\
   & 72 & 2 & 2 & 2 & 6 & 0.0 & 2 & 16 & 0.0 \\
   & 73 & 3 & 3 & 3 & 1 & 0.0 & 3 & 1 & 0.0 \\
   & 74 & 3 & 3 & 3 & 1 & 0.0 & 3 & 1 & 0.0 \\
   & 75 & 2 & 2 & 2 & 2 & 0.0 & 2 & 1 & 0.0 \\
   & 76 & 2 & 2 & 2 & 1 & 0.0 & 2 & 1 & 0.0 \\
   & 77 & 2 & 2 & 2 & 1 & 0.0 & 2 & 1 & 0.0 \\
   & 78 & 3 & 3 & 3 & 1 & 0.0 & 3 & 1 & 0.0 \\
   & 79 & 2 & 2 & 2 & 1 & 0.0 & 2 & 4 & 0.0 \\
   & 80 & 3 & 3 & 3 & 1 & 0.0 & 3 & 1 & 0.0 \\ \hline
80 & 81 & 3 & 3 & 3 & 1 & 0.0 & 3 & 1 & 0.0 \\
   & 82 & 3 & 3 & 3 & 1 & 0.0 & 3 & 1 & 0.0 \\
   & 83 & 3 & 3 & 3 & 1 & 0.0 & 3 & 1 & 0.0 \\
   & 84 & 3 & 3 & 3 & 1 & 0.0 & 3 & 1 & 0.0 \\
   & 85 & 3 & 3 & 3 & 1 & 0.0 & 3 & 1 & 0.0 \\
   & 86 & 3 & 3 & 3 & 1 & 0.0 & 3 & 1 & 0.0 \\
   & 87 & 3 & 3 & 3 & 4 & 0.0 & 3 & 13 & 0.0 \\
   & 88 & 3 & 3 & 3 & 2 & 0.0 & 3 & 1 & 0.0 \\
   & 89 & 4 & 4 & 4 & 1 & 0.0 & 4 & 1 & 0.0 \\
   & 90 & 3 & 3 & 3 & 1 & 0.0 & 3 & 1 & 0.0 \\ \hline
100 & 91 & 4 & 4 & 4 & 1 & 0.0 & 4 & 1 & 0.0 \\
   & 92 & 4 & 4 & 4 & 1 & 0.0 & 4 & 1 & 0.0 \\
   & 93 & 3 & 3 & 3 & 41 & 0.0 & 3 & 8 & 0.0 \\
   & 94 & 4 & 4 & 4 & 1 & 0.0 & 4 & 1 & 0.0 \\
   & 95 & 4 & 4 & 4 & 1 & 0.0 & 4 & 1 & 0.0 \\
   & 96 & 4 & 4 & 4 & 1 & 0.0 & 4 & 1 & 0.0 \\
   & 97 & 4 & 4 & 4 & 1 & 0.0 & 4 & 1 & 0.0 \\
   & 98 & 4 & 4 & 4 & 1 & 0.0 & 4 & 1 & 0.0 \\
   & 99 & 4 & 4 & 4 & 1 & 0.0 & 4 & 1 & 0.0 \\
   & 100 & 4 & 4 & 4 & 1 & 0.0 & 4 & 1 & 0.0 \\ \hline
\end{tabular}}
\end{table}

\begin{table}[!t]
\centering
\caption{Detailed results for the 50 problem instances of Class~III.}
\label{tab:detailed-results-class3}
\scalebox{0.8}{
\begin{tabular}{ll|ll|lll|lll} \hline
{\bf \#}    &            &          &          & \multicolumn{3}{|c|}{\bf MS-LGFi} & \multicolumn{3}{|c}{\bf EA-LGFi} \\
{\bf items} & {\bf inst} & {\bf LB} & {\bf UB} & {\bf res} & {\bf eval} & {\bf time} & {\bf res} & {\bf eval} & {\bf time} \\ \hline
20 & 101 & 6 & 6 & 6 & 1 & 0.0 & 6 & 1 & 0.0 \\
   & 102 & 3 & 3 & 3 & 3675 & 0.0 & 3 & 11741 & 0.2 \\
   & 103 & 6 & 6 & 6 & 1 & 0.0 & 6 & 1 & 0.0 \\
   & 104 & 4 & 4 & 4 & 4 & 0.0 & 4 & 3 & 0.0 \\
   & 105 & 4 & 4 & 4 & 36 & 0.0 & 4 & 49 & 0.0 \\
   & 106 & 7 & 7 & 7 & 1 & 0.0 & 7 & 1 & 0.0 \\
   & 107 & 5 & 5 & 5 & 1 & 0.0 & 5 & 1 & 0.0 \\
   & 108 & 4 & 4 & 4 & 5 & 0.0 & 4 & 7 & 0.0 \\
   & 109 & 5 & 5 & 5 & 2 & 0.0 & 5 & 3 & 0.0 \\
   & 110 & 7 & 7 & 7 & 1 & 0.0 & 7 & 1 & 0.0 \\ \hline
40 & 111 & 6 & 6 & 6 & 358 & 0.0 & 6 & 91 & 0.0 \\
   & 112 & 8 & 8 & 8 & 2327 & 0.1 & 8 & 583 & 0.0 \\
   & 113 & 11 & 11 & 11 & 3 & 0.0 & 11 & 1 & 0.0 \\
   & 114 & 10 & 10 & 10 & 14 & 0.0 & 10 & 1 & 0.0 \\
   & 115 & 12 & 12 & 12 & 1 & 0.0 & 12 & 1 & 0.0 \\
   & 116 & 10 & 10 & 10 & 1 & 0.0 & 10 & 1 & 0.0 \\
   & 117 & 8 & 8 & 8 & 7831 & 0.2 & 8 & 3549 & 0.1 \\
   & 118 & 13 & 13 & 13 & 6 & 0.0 & 13 & 38 & 0.0 \\
   & 119 & 7 & 8 & 8 & 2 & 0.0 & 8 & 1 & 0.0 \\
   & 120 & 7 & 8 & 8 & 3 & 0.0 & 8 & 4 & 0.0 \\ \hline
60 & 121 & 16 & 16 & 16 & 9 & 0.0 & 16 & 21 & 0.0 \\
   & 122 & 12 & 13 & 13 & 5 & 0.0 & 13 & 9 & 0.0 \\
   & 123 & 13 & 14 & 14 & 2 & 0.0 & 14 & 3 & 0.0 \\
   & 124 & 14 & 15 & 15 & 3 & 0.0 & 15 & 2 & 0.0 \\
   & 125 & 12 & 12 & 12 & 27 & 0.0 & 12 & 331 & 0.0 \\
   & 126 & 12 & 12 & 12 & 6170 & 0.3 & 12 & 52890 & 2.7 \\
   & 127 & 11 & 11 & 11 & 90 & 0.0 & 11 & 11 & 0.0 \\
   & 128 & 15 & 15 & 15 & 4 & 0.0 & 15 & 1 & 0.0 \\
   & 129 & 13 & 13 & 13 & 18 & 0.0 & 13 & 9 & 0.0 \\
   & 130 & 18 & 18 & 18 & 1 & 0.0 & 18 & 2 & 0.0 \\ \hline
80 & 131 & 17 & 17 & \fcolorbox{black}{white}{18} & 2 & 0.0 & 17 & 1932 & 0.2 \\
   & 132 & 18 & 18 & 18 & 3936 & 0.3 & 18 & 3902 & 0.3 \\
   & 133 & 17 & 18 & 18 & 2 & 0.0 & 18 & 7 & 0.0 \\
   & 134 & 18 & 18 & 18 & 902 & 0.1 & 18 & 153 & 0.0 \\
   & 135 & 16 & 17 & 17 & 49 & 0.0 & 17 & 57 & 0.0 \\
   & 136 & 20 & 20 & 20 & 29 & 0.0 & 20 & 2 & 0.0 \\
   & 137 & 20 & 20 & \fcolorbox{black}{white}{21} & 8 & 0.0 & 20 & 2500617 & 192.3 \\
   & 138 & 22 & 22 & 22 & 2 & 0.0 & 22 & 2 & 0.0 \\
   & 139 & 21 & 21 & 21 & 496 & 0.0 & 21 & 767 & 0.1 \\
   & 140 & 18 & 18 & 18 & 38251 & 2.7 & 18 & 184395 & 14.0 \\ \hline
100 & 141 & 19 & 19 & 19 & 6585 & 0.6 & 19 & 5573 & 0.6 \\
   & 142 & 22 & 22 & \fcolorbox{black}{white}{23} & 14 & 0.0 & \fcolorbox{black}{white}{23} & 5 & 0.0 \\
   & 143 & 18 & 19 & 19 & 19949 & 2.0 & 19 & 1406 & 0.1 \\
   & 144 & 20 & 20 & \fcolorbox{black}{white}{21} & 1 & 0.0 & 20 & 2480794 & 260.5 \\
   & 145 & 22 & 22 & 22 & 73503 & 7.3 & 22 & 2882 & 0.3 \\
   & 146 & 27 & 27 & 27 & 3 & 0.0 & 27 & 2 & 0.0 \\
   & 147 & 20 & 20 & 20 & 3159 & 0.3 & 20 & 933 & 0.1 \\
   & 148 & 23 & 23 & 23 & 761 & 0.1 & 23 & 11 & 0.0 \\
   & 149 & 21 & 22 & 22 & 107 & 0.0 & 22 & 64 & 0.0 \\
   & 150 & 29 & 29 & 29 & 1 & 0.0 & 29 & 3 & 0.0 \\ \hline
\end{tabular}}
\end{table}

\begin{table}[!t]
\centering
\caption{Detailed results for the 50 problem instances of Class~IV.}
\label{tab:detailed-results-class4}
\scalebox{0.8}{
\begin{tabular}{ll|ll|lll|lll} \hline
{\bf \#}    &            &          &          & \multicolumn{3}{|c|}{\bf MS-LGFi} & \multicolumn{3}{|c}{\bf EA-LGFi} \\
{\bf items} & {\bf inst} & {\bf LB} & {\bf UB} & {\bf res} & {\bf eval} & {\bf time} & {\bf res} & {\bf eval} & {\bf time} \\ \hline
20 & 151 & 1 & 1 & 1 & 1 & 0.0 & 1 & 1 & 0.0 \\
   & 152 & 1 & 1 & 1 & 1 & 0.0 & 1 & 1 & 0.0 \\
   & 153 & 1 & 1 & 1 & 1 & 0.0 & 1 & 1 & 0.0 \\
   & 154 & 1 & 1 & 1 & 1 & 0.0 & 1 & 1 & 0.0 \\
   & 155 & 1 & 1 & 1 & 1 & 0.0 & 1 & 1 & 0.0 \\
   & 156 & 1 & 1 & 1 & 2 & 0.0 & 1 & 1 & 0.0 \\
   & 157 & 1 & 1 & 1 & 1 & 0.0 & 1 & 1 & 0.0 \\
   & 158 & 1 & 1 & 1 & 1 & 0.0 & 1 & 1 & 0.0 \\
   & 159 & 1 & 1 & 1 & 1 & 0.0 & 1 & 1 & 0.0 \\
   & 160 & 1 & 1 & 1 & 1 & 0.0 & 1 & 1 & 0.0 \\ \hline
40 & 161 & 1 & 1 & 1 & 1 & 0.0 & 1 & 1 & 0.0 \\
   & 162 & 2 & 2 & 2 & 1 & 0.0 & 2 & 1 & 0.0 \\
   & 163 & 2 & 2 & 2 & 1 & 0.0 & 2 & 1 & 0.0 \\
   & 164 & 2 & 2 & 2 & 1 & 0.0 & 2 & 1 & 0.0 \\
   & 165 & 2 & 2 & 2 & 1 & 0.0 & 2 & 1 & 0.0 \\
   & 166 & 2 & 2 & 2 & 1 & 0.0 & 2 & 1 & 0.0 \\
   & 167 & 2 & 2 & 2 & 1 & 0.0 & 2 & 1 & 0.0 \\
   & 168 & 2 & 2 & 2 & 1 & 0.0 & 2 & 1 & 0.0 \\
   & 169 & 2 & 2 & 2 & 1 & 0.0 & 2 & 1 & 0.0 \\
   & 170 & 2 & 2 & 2 & 1 & 0.0 & 2 & 1 & 0.0 \\ \hline
60 & 171 & 3 & 3 & 3 & 1 & 0.0 & 3 & 1 & 0.0 \\
   & 172 & 2 & 2 & 2 & 3 & 0.0 & 2 & 1 & 0.0 \\
   & 173 & 2 & 3 & \fcolorbox{black}{lightgray}{2$^\ast$} & 1422800 & 71.1 & \fcolorbox{black}{lightgray}{2$^\ast$} & 1729156 & 96.4 \\
   & 174 & 2 & 3 & \fcolorbox{black}{lightgray}{2$^\ast$} & 972745 & 46.5 & \fcolorbox{black}{lightgray}{2$^\ast$} & 456033 & 25.4 \\
   & 175 & 2 & 2 & 2 & 1 & 0.0 & 2 & 1 & 0.0 \\
   & 176 & 2 & 2 & 2 & 1 & 0.0 & 2 & 1 & 0.0 \\
   & 177 & 2 & 2 & 2 & 1 & 0.0 & 2 & 1 & 0.0 \\
   & 178 & 3 & 3 & 3 & 1 & 0.0 & 3 & 1 & 0.0 \\
   & 179 & 2 & 2 & 2 & 1 & 0.0 & 2 & 2 & 0.0 \\
   & 180 & 3 & 3 & 3 & 1 & 0.0 & 3 & 1 & 0.0 \\ \hline
80 & 181 & 3 & 3 & 3 & 1 & 0.0 & 3 & 1 & 0.0 \\
   & 182 & 3 & 3 & 3 & 1 & 0.0 & 3 & 1 & 0.0 \\
   & 183 & 3 & 3 & 3 & 1 & 0.0 & 3 & 1 & 0.0 \\
   & 184 & 3 & 3 & 3 & 1 & 0.0 & 3 & 1 & 0.0 \\
   & 185 & 3 & 3 & 3 & 1 & 0.0 & 3 & 1 & 0.0 \\
   & 186 & 3 & 3 & 3 & 1 & 0.0 & 3 & 1 & 0.0 \\
   & 187 & 3 & 3 & 3 & 242 & 0.0 & 3 & 84 & 0.0 \\
   & 188 & 3 & 3 & 3 & 69 & 0.0 & 3 & 16 & 0.0 \\
   & 189 & 3 & 4 & 4 & 1 & 0.0 & 4 & 1 & 0.0 \\
   & 190 & 3 & 3 & 3 & 1 & 0.0 & 3 & 1 & 0.0 \\ \hline
100 & 191 & 3 & 3 & 3 & 6 & 0.0 & 3 & 5 & 0.0 \\
   & 192 & 4 & 4 & 4 & 1 & 0.0 & 4 & 1 & 0.0 \\
   & 193 & 3 & 3 & 3 & 1 & 0.0 & 3 & 1 & 0.0 \\
   & 194 & 4 & 4 & 4 & 1 & 0.0 & 4 & 1 & 0.0 \\
   & 195 & 4 & 4 & 4 & 1 & 0.0 & 4 & 1 & 0.0 \\
   & 196 & 4 & 4 & 4 & 1 & 0.0 & 4 & 1 & 0.0 \\
   & 197 & 3 & 4 & \fcolorbox{black}{lightgray}{3$^\ast$} & 1080 & 0.1 & \fcolorbox{black}{lightgray}{3$^\ast$} & 277 & 0.0 \\
   & 198 & 4 & 4 & 4 & 1 & 0.0 & 4 & 1 & 0.0 \\
   & 199 & 4 & 4 & 4 & 1 & 0.0 & 4 & 1 & 0.0 \\
   & 200 & 4 & 4 & 4 & 1 & 0.0 & 4 & 1 & 0.0 \\ \hline
\end{tabular}}
\end{table}

\begin{table}[!t]
\centering
\caption{Detailed results for the 50 problem instances of Class~V.}
\label{tab:detailed-results-class5}
\scalebox{0.8}{
\begin{tabular}{ll|ll|lll|lll} \hline
{\bf \#}    &            &          &          & \multicolumn{3}{|c|}{\bf MS-LGFi} & \multicolumn{3}{|c}{\bf EA-LGFi} \\
{\bf items} & {\bf inst} & {\bf LB} & {\bf UB} & {\bf res} & {\bf eval} & {\bf time} & {\bf res} & {\bf eval} & {\bf time} \\ \hline
20 & 201 & 8 & 8 & 8 & 1 & 0.0 & 8 & 1 & 0.0 \\
   & 202 & 5 & 5 & 5 & 1 & 0.0 & 5 & 1 & 0.0 \\
   & 203 & 7 & 7 & 7 & 1 & 0.0 & 7 & 3 & 0.0 \\
   & 204 & 5 & 5 & 5 & 1 & 0.0 & 5 & 1 & 0.0 \\
   & 205 & 5 & 5 & 5 & 341 & 0.0 & 5 & 1268 & 0.0 \\
   & 206 & 9 & 9 & 9 & 1 & 0.0 & 9 & 1 & 0.0 \\
   & 207 & 6 & 6 & 6 & 1 & 0.0 & 6 & 1 & 0.0 \\
   & 208 & 5 & 5 & 5 & 1 & 0.0 & 5 & 1 & 0.0 \\
   & 209 & 7 & 7 & 7 & 1 & 0.0 & 7 & 1 & 0.0 \\
   & 210 & 8 & 8 & 8 & 1 & 0.0 & 8 & 1 & 0.0 \\ \hline
40 & 211 & 8 & 8 & 8 & 92 & 0.0 & 8 & 29 & 0.0 \\
   & 212 & 10 & 10 & 10 & 30 & 0.0 & 10 & 117 & 0.0 \\
   & 213 & 15 & 15 & 15 & 1 & 0.0 & 15 & 1 & 0.0 \\
   & 214 & 13 & 13 & 13 & 1 & 0.0 & 13 & 1 & 0.0 \\
   & 215 & 14 & 14 & 14 & 1 & 0.0 & 14 & 1 & 0.0 \\
   & 216 & 12 & 12 & 12 & 1 & 0.0 & 12 & 1 & 0.0 \\
   & 217 & 10 & 10 & 10 & 63 & 0.0 & 10 & 103 & 0.0 \\
   & 218 & 17 & 17 & 17 & 1 & 0.0 & 17 & 1 & 0.0 \\
   & 219 & 10 & 10 & 10 & 1 & 0.0 & 10 & 1 & 0.0 \\
   & 220 & 10 & 10 & 10 & 19332 & 0.5 & 10 & 8046 & 0.3 \\ \hline
60 & 221 & 20 & 20 & 20 & 10303 & 0.6 & 20 & 8882 & 0.5 \\
   & 222 & 17 & 17 & 17 & 1 & 0.0 & 17 & 1 & 0.0 \\
   & 223 & 19 & 19 & 19 & 2 & 0.0 & 19 & 5 & 0.0 \\
   & 224 & 20 & 20 & 20 & 1 & 0.0 & 20 & 1 & 0.0 \\
   & 225 & 15 & 15 & 15 & 384 & 0.0 & 15 & 433 & 0.0 \\
   & 226 & 15 & 16 & 16 & 2 & 0.0 & 16 & 1 & 0.0 \\
   & 227 & 14 & 14 & 14 & 5 & 0.0 & 14 & 21 & 0.0 \\
   & 228 & 19 & 19 & 19 & 2 & 0.0 & 19 & 11 & 0.0 \\
   & 229 & 16 & 16 & 16 & 1722 & 0.1 & 16 & 16211 & 0.9 \\
   & 230 & 24 & 24 & 24 & 1 & 0.0 & 24 & 1 & 0.0 \\ \hline
80 & 231 & 22 & 22 & 22 & 102 & 0.0 & 22 & 101 & 0.0 \\
   & 232 & 22 & 23 & 23 & 13 & 0.0 & 23 & 10 & 0.0 \\
   & 233 & 24 & 25 & 25 & 1 & 0.0 & 25 & 2 & 0.0 \\
   & 234 & 25 & 25 & 25 & 1 & 0.0 & 25 & 1 & 0.0 \\
   & 235 & 22 & 23 & 23 & 1 & 0.0 & 23 & 14 & 0.0 \\
   & 236 & 25 & 26 & 26 & 1 & 0.0 & 26 & 1 & 0.0 \\
   & 237 & 27 & 27 & 27 & 11 & 0.0 & 27 & 29 & 0.0 \\
   & 238 & 26 & 26 & 26 & 11164 & 0.9 & 26 & 3376 & 0.3 \\
   & 239 & 26 & 27 & 27 & 226 & 0.0 & 27 & 19 & 0.0 \\
   & 240 & 22 & 23 & 23 & 244 & 0.0 & 23 & 447 & 0.0 \\ \hline
100 & 241 & 23 & 24 & \fcolorbox{black}{white}{25} & 5 & 0.0 & \fcolorbox{black}{white}{25} & 8 & 0.0 \\
   & 242 & 28 & 29 & 29 & 28 & 0.0 & 29 & 12 & 0.0 \\
   & 243 & 24 & 24 & 24 & 135288 & 15.3 & 24 & 632 & 0.1 \\
   & 244 & 26 & 26 & \fcolorbox{black}{white}{27} & 67 & 0.0 & 26 & 1309655 & 152.7 \\
   & 245 & 28 & 28 & \fcolorbox{black}{white}{29} & 68 & 0.0 & 28 & 834082 & 93.8 \\
   & 246 & 34 & 34 & 34 & 1 & 0.0 & 34 & 2 & 0.0 \\
   & 247 & 25 & 25 & \fcolorbox{black}{white}{26} & 1 & 0.0 & 25 & 236926 & 26.7 \\
   & 248 & 29 & 29 & \fcolorbox{black}{white}{30} & 265 & 0.0 & \fcolorbox{black}{white}{30} & 106 & 0.0 \\
   & 249 & 27 & 27 & \fcolorbox{black}{white}{28} & 42 & 0.0 & \fcolorbox{black}{white}{28} & 61 & 0.0 \\
   & 250 & 35 & 35 & 35 & 4 & 0.0 & 35 & 1 & 0.0 \\ \hline
\end{tabular}}
\end{table}

\begin{table}[!t]
\centering
\caption{Detailed results for the 50 problem instances of Class~VI.}
\label{tab:detailed-results-class6}
\scalebox{0.8}{
\begin{tabular}{ll|ll|lll|lll} \hline
{\bf \#}    &            &          &          & \multicolumn{3}{|c|}{\bf MS-LGFi} & \multicolumn{3}{|c}{\bf EA-LGFi} \\
{\bf items} & {\bf inst} & {\bf LB} & {\bf UB} & {\bf res} & {\bf eval} & {\bf time} & {\bf res} & {\bf eval} & {\bf time} \\ \hline
20 & 251 & 1 & 1 & 1 & 1 & 0.0 & 1 & 1 & 0.0 \\
   & 252 & 1 & 1 & 1 & 1 & 0.0 & 1 & 1 & 0.0 \\
   & 253 & 1 & 1 & 1 & 1 & 0.0 & 1 & 1 & 0.0 \\
   & 254 & 1 & 1 & 1 & 1 & 0.0 & 1 & 1 & 0.0 \\
   & 255 & 1 & 1 & 1 & 1 & 0.0 & 1 & 1 & 0.0 \\
   & 256 & 1 & 1 & 1 & 1 & 0.0 & 1 & 1 & 0.0 \\
   & 257 & 1 & 1 & 1 & 1 & 0.0 & 1 & 1 & 0.0 \\
   & 258 & 1 & 1 & 1 & 1 & 0.0 & 1 & 1 & 0.0 \\
   & 259 & 1 & 1 & 1 & 1 & 0.0 & 1 & 1 & 0.0 \\
   & 260 & 1 & 1 & 1 & 1 & 0.0 & 1 & 1 & 0.0 \\ \hline
40 & 261 & 1 & 1 & 1 & 1 & 0.0 & 1 & 1 & 0.0 \\
   & 262 & 1 & 2 & 2 & 1 & 0.0 & 2 & 1 & 0.0 \\
   & 263 & 2 & 2 & 2 & 1 & 0.0 & 2 & 1 & 0.0 \\
   & 264 & 2 & 2 & 2 & 1 & 0.0 & 2 & 1 & 0.0 \\
   & 265 & 2 & 2 & 2 & 1 & 0.0 & 2 & 1 & 0.0 \\
   & 266 & 1 & 2 & 2 & 1 & 0.0 & 2 & 1 & 0.0 \\
   & 267 & 2 & 2 & 2 & 1 & 0.0 & 2 & 1 & 0.0 \\
   & 268 & 2 & 2 & 2 & 1 & 0.0 & 2 & 1 & 0.0 \\
   & 269 & 1 & 1 & 1 & 114 & 0.0 & 1 & 96 & 0.0 \\
   & 270 & 1 & 1 & 1 & 3013 & 0.1 & 1 & 6880 & 0.3 \\ \hline
60 & 271 & 2 & 2 & 2 & 272 & 0.0 & 2 & 112 & 0.0 \\
   & 272 & 2 & 2 & 2 & 1 & 0.0 & 2 & 1 & 0.0 \\
   & 273 & 2 & 2 & 2 & 1 & 0.0 & 2 & 1 & 0.0 \\
   & 274 & 2 & 2 & 2 & 1 & 0.0 & 2 & 1 & 0.0 \\
   & 275 & 2 & 2 & 2 & 1 & 0.0 & 2 & 1 & 0.0 \\
   & 276 & 2 & 2 & 2 & 1 & 0.0 & 2 & 1 & 0.0 \\
   & 277 & 2 & 2 & 2 & 1 & 0.0 & 2 & 1 & 0.0 \\
   & 278 & 2 & 2 & 2 & 1 & 0.0 & 2 & 1 & 0.0 \\
   & 279 & 2 & 2 & 2 & 1 & 0.0 & 2 & 1 & 0.0 \\
   & 280 & 3 & 3 & 3 & 1 & 0.0 & 3 & 1 & 0.0 \\ \hline
80 & 281 & 3 & 3 & 3 & 1 & 0.0 & 3 & 1 & 0.0 \\
   & 282 & 3 & 3 & 3 & 1 & 0.0 & 3 & 1 & 0.0 \\
   & 283 & 3 & 3 & 3 & 1 & 0.0 & 3 & 1 & 0.0 \\
   & 284 & 3 & 3 & 3 & 1 & 0.0 & 3 & 1 & 0.0 \\
   & 285 & 3 & 3 & 3 & 1 & 0.0 & 3 & 1 & 0.0 \\
   & 286 & 3 & 3 & 3 & 1 & 0.0 & 3 & 1 & 0.0 \\
   & 287 & 3 & 3 & 3 & 1 & 0.0 & 3 & 1 & 0.0 \\
   & 288 & 3 & 3 & 3 & 1 & 0.0 & 3 & 1 & 0.0 \\
   & 289 & 3 & 3 & 3 & 1 & 0.0 & 3 & 1 & 0.0 \\
   & 290 & 3 & 3 & 3 & 1 & 0.0 & 3 & 1 & 0.0 \\ \hline
100 & 291 & 3 & 3 & 3 & 1 & 0.0 & 3 & 1 & 0.0 \\
   & 292 & 3 & 3 & 3 & 13896 & 1.5 & 3 & 37796 & 4.6 \\
   & 293 & 3 & 3 & 3 & 1 & 0.0 & 3 & 1 & 0.0 \\
   & 294 & 3 & 3 & 3 & 1 & 0.0 & 3 & 1 & 0.0 \\
   & 295 & 3 & 3 & 3 & 2 & 0.0 & 3 & 3 & 0.0 \\
   & 296 & 4 & 4 & 4 & 1 & 0.0 & 4 & 1 & 0.0 \\
   & 297 & 3 & 3 & 3 & 1 & 0.0 & 3 & 1 & 0.0 \\
   & 298 & 3 & 4 & \fcolorbox{black}{lightgray}{3$^\ast$} & 2615111 & 294.2 & \fcolorbox{black}{lightgray}{3$^\ast$} & 9555 & 1.2 \\
   & 299 & 3 & 3 & 3 & 2 & 0.0 & 3 & 1 & 0.0 \\
   & 300 & 4 & 4 & 4 & 1 & 0.0 & 4 & 1 & 0.0 \\ \hline
\end{tabular}}
\end{table}

\begin{table}[!t]
\centering
\caption{Detailed results for the 50 problem instances of Class~VII.}
\label{tab:detailed-results-class7}
\scalebox{0.8}{
\begin{tabular}{ll|ll|lll|lll} \hline
{\bf \#}    &            &          &          & \multicolumn{3}{|c|}{\bf MS-LGFi} & \multicolumn{3}{|c}{\bf EA-LGFi} \\
{\bf items} & {\bf inst} & {\bf LB} & {\bf UB} & {\bf res} & {\bf eval} & {\bf time} & {\bf res} & {\bf eval} & {\bf time} \\ \hline
20 & 301 & 5 & 5 & 5 & 3 & 0.0 & 5 & 6 & 0.0 \\
   & 302 & 5 & 5 & 5 & 1 & 0.0 & 5 & 1 & 0.0 \\
   & 303 & 5 & 5 & 5 & 4 & 0.0 & 5 & 4 & 0.0 \\
   & 304 & 7 & 7 & 7 & 1 & 0.0 & 7 & 1 & 0.0 \\
   & 305 & 6 & 6 & 6 & 1 & 0.0 & 6 & 1 & 0.0 \\
   & 306 & 6 & 6 & 6 & 1 & 0.0 & 6 & 1 & 0.0 \\
   & 307 & 4 & 4 & 4 & 1 & 0.0 & 4 & 2 & 0.0 \\
   & 308 & 7 & 7 & 7 & 1 & 0.0 & 7 & 1 & 0.0 \\
   & 309 & 6 & 6 & 6 & 21 & 0.0 & 6 & 32 & 0.0 \\
   & 310 & 4 & 4 & 4 & 2 & 0.0 & 4 & 4 & 0.0 \\ \hline
40 & 311 & 10 & 10 & 10 & 1 & 0.0 & 10 & 1 & 0.0 \\
   & 312 & 12 & 12 & 12 & 71 & 0.0 & 12 & 1108 & 0.0 \\
   & 313 & 9 & 10 & 10 & 1 & 0.0 & 10 & 1 & 0.0 \\
   & 314 & 14 & 14 & 14 & 7 & 0.0 & 14 & 2 & 0.0 \\
   & 315 & 10 & 10 & 10 & 1 & 0.0 & 10 & 1 & 0.0 \\
   & 316 & 11 & 11 & 11 & 42 & 0.0 & 11 & 29 & 0.0 \\
   & 317 & 11 & 12 & 12 & 1 & 0.0 & 12 & 1 & 0.0 \\
   & 318 & 11 & 11 & 11 & 5645 & 0.1 & 11 & 1931 & 0.1 \\
   & 319 & 8 & 8 & 8 & 5 & 0.0 & 8 & 1 & 0.0 \\
   & 320 & 13 & 13 & 13 & 1 & 0.0 & 13 & 1 & 0.0 \\ \hline
60 & 321 & 17 & 17 & \fcolorbox{black}{white}{18} & 1 & 0.0 & \fcolorbox{black}{white}{18} & 1 & 0.0 \\
   & 322 & 14 & 14 & 14 & 1 & 0.0 & 14 & 1 & 0.0 \\
   & 323 & 17 & 17 & 17 & 1 & 0.0 & 17 & 1 & 0.0 \\
   & 324 & 15 & 15 & 15 & 1 & 0.0 & 15 & 2 & 0.0 \\
   & 325 & 14 & 15 & 15 & 1 & 0.0 & 15 & 1 & 0.0 \\
   & 326 & 15 & 15 & 15 & 1 & 0.0 & 15 & 2 & 0.0 \\
   & 327 & 15 & 15 & 15 & 1 & 0.0 & 15 & 1 & 0.0 \\
   & 328 & 17 & 17 & 17 & 141 & 0.0 & 17 & 1 & 0.0 \\
   & 329 & 14 & 14 & 14 & 4 & 0.0 & 14 & 57 & 0.0 \\
   & 330 & 18 & 19 & 19 & 1 & 0.0 & 19 & 5 & 0.0 \\ \hline
80 & 331 & 20 & 21 & 21 & 5 & 0.0 & 21 & 6 & 0.0 \\
   & 332 & 25 & 25 & 25 & 1 & 0.0 & 25 & 3 & 0.0 \\
   & 333 & 20 & 21 & 21 & 1 & 0.0 & 21 & 1 & 0.0 \\
   & 334 & 21 & 22 & 22 & 4 & 0.0 & 22 & 16 & 0.0 \\
   & 335 & 23 & 24 & 24 & 1 & 0.0 & 24 & 1 & 0.0 \\
   & 336 & 22 & 23 & 23 & 1 & 0.0 & 23 & 1 & 0.0 \\
   & 337 & 24 & 25 & 25 & 1 & 0.0 & 25 & 1 & 0.0 \\
   & 338 & 22 & 23 & 23 & 1 & 0.0 & 23 & 1 & 0.0 \\
   & 339 & 23 & 24 & 24 & 2 & 0.0 & 24 & 1 & 0.0 \\
   & 340 & 24 & 24 & 24 & 2 & 0.0 & 24 & 4 & 0.0 \\ \hline
100 & 341 & 27 & 27 & 27 & 248 & 0.0 & 27 & 52 & 0.0 \\
   & 342 & 27 & 27 & 27 & 1 & 0.0 & 27 & 14 & 0.0 \\
   & 343 & 24 & 25 & 25 & 1 & 0.0 & 25 & 1 & 0.0 \\
   & 344 & 26 & 27 & 27 & 1 & 0.0 & 27 & 1 & 0.0 \\
   & 345 & 25 & 25 & 25 & 1 & 0.0 & 25 & 1 & 0.0 \\
   & 346 & 28 & 28 & 28 & 2791 & 0.3 & 28 & 413 & 0.0 \\
   & 347 & 27 & 27 & 27 & 1 & 0.0 & 27 & 1 & 0.0 \\
   & 348 & 29 & 29 & 29 & 2 & 0.0 & 29 & 1 & 0.0 \\
   & 349 & 25 & 25 & 25 & 18 & 0.0 & 25 & 32 & 0.0 \\
   & 350 & 31 & 31 & 31 & 52488 & 5.0 & 31 & 501 & 0.1 \\ \hline
\end{tabular}}
\end{table}

\begin{table}[!t]
\centering
\caption{Detailed results for the 50 problem instances of Class~VIII.}
\label{tab:detailed-results-class8}
\scalebox{0.8}{
\begin{tabular}{ll|ll|lll|lll} \hline
{\bf \#}    &            &          &          & \multicolumn{3}{|c|}{\bf MS-LGFi} & \multicolumn{3}{|c}{\bf EA-LGFi} \\
{\bf items} & {\bf inst} & {\bf LB} & {\bf UB} & {\bf res} & {\bf eval} & {\bf time} & {\bf res} & {\bf eval} & {\bf time} \\ \hline
20 & 351 & 6 & 6 & 6 & 2 & 0.0 & 6 & 1 & 0.0 \\
   & 352 & 7 & 7 & 7 & 1 & 0.0 & 7 & 1 & 0.0 \\
   & 353 & 5 & 5 & 5 & 14387 & 0.1 & 5 & 20148 & 0.3 \\
   & 354 & 7 & 7 & 7 & 1 & 0.0 & 7 & 2 & 0.0 \\
   & 355 & 6 & 6 & 6 & 1 & 0.0 & 6 & 1 & 0.0 \\
   & 356 & 6 & 6 & 6 & 1 & 0.0 & 6 & 1 & 0.0 \\
   & 357 & 5 & 5 & 5 & 1 & 0.0 & 5 & 1 & 0.0 \\
   & 358 & 5 & 5 & 5 & 22 & 0.0 & 5 & 32 & 0.0 \\
   & 359 & 7 & 7 & 7 & 1 & 0.0 & 7 & 1 & 0.0 \\
   & 360 & 4 & 4 & 4 & 1 & 0.0 & 4 & 2 & 0.0 \\ \hline
40 & 361 & 11 & 12 & 12 & 1 & 0.0 & 12 & 2 & 0.0 \\
   & 362 & 13 & 13 & 13 & 26 & 0.0 & 13 & 403 & 0.0 \\
   & 363 & 11 & 11 & 11 & 2 & 0.0 & 11 & 1 & 0.0 \\
   & 364 & 12 & 12 & 12 & 1 & 0.0 & 12 & 1 & 0.0 \\
   & 365 & 9 & 9 & 9 & 1 & 0.0 & 9 & 2 & 0.0 \\
   & 366 & 12 & 12 & 12 & 4 & 0.0 & 12 & 2 & 0.0 \\
   & 367 & 11 & 11 & 11 & 3662 & 0.1 & 11 & 452 & 0.0 \\
   & 368 & 11 & 11 & 11 & 4 & 0.0 & 11 & 10 & 0.0 \\
   & 369 & 9 & 9 & 9 & 2 & 0.0 & 9 & 4 & 0.0 \\
   & 370 & 13 & 13 & 13 & 10 & 0.0 & 13 & 2 & 0.0 \\ \hline
60 & 371 & 17 & 17 & 17 & 10 & 0.0 & 17 & 29 & 0.0 \\
   & 372 & 17 & 17 & 17 & 1 & 0.0 & 17 & 1 & 0.0 \\
   & 373 & 16 & 17 & 17 & 4 & 0.0 & 17 & 2 & 0.0 \\
   & 374 & 15 & 15 & 15 & 41 & 0.0 & 15 & 78 & 0.0 \\
   & 375 & 14 & 14 & 14 & 4 & 0.0 & 14 & 2 & 0.0 \\
   & 376 & 15 & 15 & 15 & 1 & 0.0 & 15 & 3 & 0.0 \\
   & 377 & 14 & 14 & 14 & 4 & 0.0 & 14 & 1 & 0.0 \\
   & 378 & 17 & 17 & 17 & 339 & 0.0 & 17 & 2890 & 0.2 \\
   & 379 & 16 & 17 & 17 & 1 & 0.0 & 17 & 1 & 0.0 \\
   & 380 & 18 & 18 & 18 & 2 & 0.0 & 18 & 3 & 0.0 \\ \hline
80 & 381 & 22 & 22 & 22 & 4 & 0.0 & 22 & 4 & 0.0 \\
   & 382 & 24 & 24 & 24 & 1 & 0.0 & 24 & 1 & 0.0 \\
   & 383 & 20 & 21 & 21 & 1 & 0.0 & 21 & 1 & 0.0 \\
   & 384 & 20 & 20 & 20 & 1 & 0.0 & 20 & 1 & 0.0 \\
   & 385 & 26 & 26 & 26 & 3 & 0.0 & 26 & 3 & 0.0 \\
   & 386 & 22 & 22 & 22 & 1 & 0.0 & 22 & 1 & 0.0 \\
   & 387 & 22 & 22 & 22 & 1 & 0.0 & 22 & 1 & 0.0 \\
   & 388 & 22 & 22 & 22 & 60 & 0.0 & 22 & 5 & 0.0 \\
   & 389 & 21 & 21 & 21 & 1184 & 0.1 & 21 & 153 & 0.0 \\
   & 390 & 24 & 24 & 24 & 12 & 0.0 & 24 & 76 & 0.0 \\ \hline
100 & 391 & 26 & 27 & 27 & 1 & 0.0 & 27 & 1 & 0.0 \\
   & 392 & 27 & 27 & 27 & 15 & 0.0 & 27 & 14 & 0.0 \\
   & 393 & 24 & 24 & 24 & 1 & 0.0 & 24 & 1 & 0.0 \\
   & 394 & 30 & 30 & 31 & 8 & 0.0 & 30 & 22536 & 2.5 \\
   & 395 & 29 & 29 & 29 & 28 & 0.0 & 29 & 14 & 0.0 \\
   & 396 & 26 & 27 & 27 & 1 & 0.0 & 27 & 1 & 0.0 \\
   & 397 & 25 & 26 & 26 & 3 & 0.0 & 26 & 34 & 0.0 \\
   & 398 & 28 & 28 & 28 & 4815 & 0.5 & 28 & 61 & 0.0 \\
   & 399 & 27 & 27 & 27 & 2 & 0.0 & 27 & 2 & 0.0 \\
   & 400 & 32 & 32 & 32 & 17 & 0.0 & 32 & 2 & 0.0 \\ \hline
\end{tabular}}
\end{table}

\begin{table}[!t]
\centering
\caption{Detailed results for the 50 problem instances of Class~IX.}
\label{tab:detailed-results-class9}
\scalebox{0.8}{
\begin{tabular}{ll|ll|lll|lll} \hline
{\bf \#}    &            &          &          & \multicolumn{3}{|c|}{\bf MS-LGFi} & \multicolumn{3}{|c}{\bf EA-LGFi} \\
{\bf items} & {\bf inst} & {\bf LB} & {\bf UB} & {\bf res} & {\bf eval} & {\bf time} & {\bf res} & {\bf eval} & {\bf time} \\ \hline
20 & 401 & 19 & 19 & 19 & 1 & 0.0 & 19 & 1 & 0.0 \\
   & 402 & 13 & 13 & 13 & 1 & 0.0 & 13 & 1 & 0.0 \\
   & 403 & 14 & 14 & 14 & 2 & 0.0 & 14 & 1 & 0.0 \\
   & 404 & 16 & 16 & 16 & 1 & 0.0 & 16 & 1 & 0.0 \\
   & 405 & 16 & 16 & 16 & 1 & 0.0 & 16 & 1 & 0.0 \\
   & 406 & 14 & 14 & 14 & 1 & 0.0 & 14 & 1 & 0.0 \\
   & 407 & 9 & 9 & 9 & 1 & 0.0 & 9 & 1 & 0.0 \\
   & 408 & 14 & 14 & 14 & 1 & 0.0 & 14 & 1 & 0.0 \\
   & 409 & 15 & 15 & 15 & 1 & 0.0 & 15 & 1 & 0.0 \\
   & 410 & 13 & 13 & 13 & 1 & 0.0 & 13 & 1 & 0.0 \\ \hline
40 & 411 & 25 & 25 & 25 & 1 & 0.0 & 25 & 1 & 0.0 \\
   & 412 & 32 & 32 & 32 & 1 & 0.0 & 32 & 1 & 0.0 \\
   & 413 & 29 & 29 & 29 & 1 & 0.0 & 29 & 1 & 0.0 \\
   & 414 & 31 & 31 & 31 & 1 & 0.0 & 31 & 1 & 0.0 \\
   & 415 & 27 & 27 & 27 & 1 & 0.0 & 27 & 1 & 0.0 \\
   & 416 & 29 & 29 & 29 & 2 & 0.0 & 29 & 1 & 0.0 \\
   & 417 & 24 & 24 & 24 & 1 & 0.0 & 24 & 1 & 0.0 \\
   & 418 & 26 & 26 & 26 & 1 & 0.0 & 26 & 1 & 0.0 \\
   & 419 & 21 & 21 & 21 & 1 & 0.0 & 21 & 1 & 0.0 \\
   & 420 & 34 & 34 & 34 & 1 & 0.0 & 34 & 1 & 0.0 \\ \hline
60 & 421 & 46 & 46 & 46 & 14 & 0.0 & 46 & 19 & 0.0 \\
   & 422 & 45 & 45 & 45 & 1 & 0.0 & 45 & 1 & 0.0 \\
   & 423 & 46 & 46 & 46 & 1 & 0.0 & 46 & 1 & 0.0 \\
   & 424 & 44 & 44 & 44 & 1 & 0.0 & 44 & 1 & 0.0 \\
   & 425 & 41 & 41 & 41 & 1 & 0.0 & 41 & 1 & 0.0 \\
   & 426 & 37 & 37 & 37 & 1 & 0.0 & 37 & 1 & 0.0 \\
   & 427 & 41 & 41 & 41 & 1 & 0.0 & 41 & 1 & 0.0 \\
   & 428 & 47 & 47 & 47 & 1 & 0.0 & 47 & 1 & 0.0 \\
   & 429 & 45 & 45 & 45 & 1 & 0.0 & 45 & 1 & 0.0 \\
   & 430 & 45 & 45 & 45 & 1 & 0.0 & 45 & 1 & 0.0 \\ \hline
80 & 431 & 59 & 59 & 59 & 1 & 0.0 & 59 & 1 & 0.0 \\
   & 432 & 58 & 58 & 58 & 1 & 0.0 & 58 & 1 & 0.0 \\
   & 433 & 57 & 57 & 57 & 1 & 0.0 & 57 & 1 & 0.0 \\
   & 434 & 53 & 53 & 53 & 1 & 0.0 & 53 & 1 & 0.0 \\
   & 435 & 62 & 62 & 62 & 1 & 0.0 & 62 & 1 & 0.0 \\
   & 436 & 62 & 62 & 62 & 1 & 0.0 & 62 & 1 & 0.0 \\
   & 437 & 59 & 59 & 59 & 1 & 0.0 & 59 & 1 & 0.0 \\
   & 438 & 58 & 58 & 58 & 1 & 0.0 & 58 & 1 & 0.0 \\
   & 439 & 49 & 49 & 49 & 1 & 0.0 & 49 & 1 & 0.0 \\
   & 440 & 60 & 60 & 60 & 1 & 0.0 & 60 & 1 & 0.0 \\ \hline
100 & 441 & 71 & 71 & 71 & 17 & 0.0 & 71 & 3 & 0.0 \\
   & 442 & 64 & 64 & 64 & 1 & 0.0 & 64 & 1 & 0.0 \\
   & 443 & 68 & 68 & 68 & 1 & 0.0 & 68 & 1 & 0.0 \\
   & 444 & 78 & 78 & 78 & 1 & 0.0 & 78 & 1 & 0.0 \\
   & 445 & 65 & 65 & 65 & 1 & 0.0 & 65 & 1 & 0.0 \\
   & 446 & 71 & 71 & 71 & 1 & 0.0 & 71 & 1 & 0.0 \\
   & 447 & 66 & 66 & 66 & 1 & 0.0 & 66 & 1 & 0.0 \\
   & 448 & 74 & 74 & 74 & 1 & 0.0 & 74 & 1 & 0.0 \\
   & 449 & 66 & 66 & 66 & 1 & 0.0 & 66 & 1 & 0.0 \\
   & 450 & 72 & 72 & 72 & 1 & 0.0 & 72 & 1 & 0.0 \\ \hline
\end{tabular}}
\end{table}

\begin{table}[!t]
\centering
\caption{Detailed results for the 50 problem instances of Class~X.}
\label{tab:detailed-results-class10}
\scalebox{0.8}{
\begin{tabular}{ll|ll|lll|lll} \hline
{\bf \#}    &            &          &          & \multicolumn{3}{|c|}{\bf MS-LGFi} & \multicolumn{3}{|c}{\bf EA-LGFi} \\
{\bf items} & {\bf inst} & {\bf LB} & {\bf UB} & {\bf res} & {\bf eval} & {\bf time} & {\bf res} & {\bf eval} & {\bf time} \\ \hline
20 & 451 & 6 & 6 & 6 & 1 & 0.0 & 6 & 1 & 0.0 \\
   & 452 & 3 & 3 & 3 & 2 & 0.0 & 3 & 2 & 0.0 \\
   & 453 & 4 & 4 & 4 & 6 & 0.0 & 4 & 4 & 0.0 \\
   & 454 & 5 & 5 & 5 & 1 & 0.0 & 5 & 1 & 0.0 \\
   & 455 & 4 & 4 & 4 & 1 & 0.0 & 4 & 1 & 0.0 \\
   & 456 & 4 & 4 & 4 & 1 & 0.0 & 4 & 1 & 0.0 \\
   & 457 & 5 & 5 & 5 & 1 & 0.0 & 5 & 1 & 0.0 \\
   & 458 & 3 & 3 & 3 & 1 & 0.0 & 3 & 1 & 0.0 \\
   & 459 & 5 & 5 & 5 & 41546 & 0.5 & 5 & 15244 & 0.2 \\
   & 460 & 3 & 3 & 3 & 1 & 0.0 & 3 & 1 & 0.0 \\ \hline
40 & 461 & 8 & 8 & 8 & 1 & 0.0 & 8 & 1 & 0.0 \\
   & 462 & 8 & 8 & 8 & 1 & 0.0 & 8 & 1 & 0.0 \\
   & 463 & 9 & 9 & 9 & 5 & 0.0 & 9 & 7 & 0.0 \\
   & 464 & 6 & 6 & 6 & 1 & 0.0 & 6 & 3 & 0.0 \\
   & 465 & 6 & 6 & 6 & 1 & 0.0 & 6 & 1 & 0.0 \\
   & 466 & 6 & 6 & 6 & 1 & 0.0 & 6 & 1 & 0.0 \\
   & 467 & 7 & 7 & 7 & 1 & 0.0 & 7 & 2 & 0.0 \\
   & 468 & 7 & 7 & 7 & 2 & 0.0 & 7 & 1 & 0.0 \\
   & 469 & 8 & 8 & 8 & 2 & 0.0 & 8 & 2 & 0.0 \\
   & 470 & 9 & 9 & 9 & 1 & 0.0 & 9 & 2 & 0.0 \\ \hline
60 & 471 & 11 & 12 & 12 & 1 & 0.0 & 12 & 2 & 0.0 \\
   & 472 & 12 & 12 & 12 & 23812 & 1.1 & 12 & 6833 & 0.4 \\
   & 473 & 11 & 11 & 11 & 523309 & 24.9 & 11 & 121233 & 6.7 \\
   & 474 & 7 & 8 & 8 & 1 & 0.0 & 8 & 1 & 0.0 \\
   & 475 & 8 & 8 & 8 & 4 & 0.0 & 8 & 2 & 0.0 \\
   & 476 & 13 & 13 & 13 & 5 & 0.0 & 13 & 1 & 0.0 \\
   & 477 & 10 & 10 & 10 & 2 & 0.0 & 10 & 1 & 0.0 \\
   & 478 & 10 & 10 & 10 & 33 & 0.0 & 10 & 3 & 0.0 \\
   & 479 & 8 & 9 & 9 & 3 & 0.0 & 9 & 1 & 0.0 \\
   & 480 & 8 & 8 & 8 & 49 & 0.0 & 8 & 11 & 0.0 \\ \hline
80 & 481 & 12 & 13 & 13 & 4 & 0.0 & 13 & 5 & 0.0 \\
   & 482 & 10 & 11 & 11 & 3 & 0.0 & 11 & 2 & 0.0 \\
   & 483 & 11 & 11 & 11 & 1 & 0.0 & 11 & 4 & 0.0 \\
   & 484 & 13 & 14 & 14 & 18 & 0.0 & 14 & 11 & 0.0 \\
   & 485 & 14 & 14 & 14 & 2 & 0.0 & 14 & 8 & 0.0 \\
   & 486 & 13 & 13 & 13 & 430266 & 31.5 & 13 & 4257 & 0.3 \\
   & 487 & 14 & 14 & \fcolorbox{black}{white}{15} & 1 & 0.0 & 14 & 3082 & 0.2 \\
   & 488 & 10 & 10 & 10 & 1 & 0.0 & 10 & 10 & 0.0 \\
   & 489 & 12 & 13 & 13 & 2 & 0.0 & 13 & 2 & 0.0 \\
   & 490 & 14 & 15 & 15 & 2 & 0.0 & 15 & 3 & 0.0 \\ \hline
100 & 491 & 14 & 15 & 15 & 1 & 0.0 & 15 & 1 & 0.0 \\
   & 492 & 15 & 16 & 16 & 1 & 0.0 & 16 & 5 & 0.0 \\
   & 493 & 15 & 16 & 16 & 34 & 0.0 & 16 & 24 & 0.0 \\
   & 494 & 17 & 18 & 18 & 1 & 0.0 & 18 & 6 & 0.0 \\
   & 495 & 17 & 18 & 18 & 2 & 0.0 & 18 & 1 & 0.0 \\
   & 496 & 13 & 13 & 13 & 183 & 0.0 & 13 & 172 & 0.0 \\
   & 497 & 13 & 14 & 14 & 1 & 0.0 & 14 & 1 & 0.0 \\
   & 498 & 18 & 18 & \fcolorbox{black}{white}{19} & 1 & 0.0 & \fcolorbox{black}{white}{19} & 1 & 0.0 \\
   & 499 & 16 & 16 & 16 & 13486 & 1.3 & 16 & 6475 & 0.7 \\
   & 500 & 15 & 15 & 15 & 254 & 0.0 & 15 & 481 & 0.1 \\ \hline
\end{tabular}}
\end{table}

\end{document}